\documentclass[runningheads]{llncs}

\usepackage{eccv}

\usepackage{eccvabbrv}
\usepackage{esvect}
\usepackage{stfloats}
\usepackage{multirow}
\usepackage{algorithm}
\usepackage[noend]{algpseudocode}
\usepackage{makecell}
\usepackage{pifont}

\usepackage{graphicx}
\usepackage{booktabs}

\usepackage{threeparttable}

\newsavebox{\mytablebox}

\usepackage[table]{xcolor}

\usepackage[breaklinks,colorlinks,citecolor=eccvblue]{hyperref}

\title{Unordered Landmark Visual Navigation}

\author{
Hao Ren\textsuperscript{1}, Junzhe Zhu\textsuperscript{1}, Yihan Li\textsuperscript{1}, Zetong Bi\textsuperscript{1}, Le Zheng\textsuperscript{1}, Zhi Li\textsuperscript{1}, \\Yiqing Yuan\textsuperscript{1}, Zhaoliang Wan\textsuperscript{2}, Dizhe Zhang\textsuperscript{2}, Lu Qi\textsuperscript{2}, Hui Cheng\textsuperscript{1}\thanks{Corresponding author: Hui Cheng (chengh9@mail.sysu.edu.cn).}\\
}
\authorrunning{H. Ren et al.}

\institute{Sun Yat-sen University, Guangzhou, China \and
Insta360 Research, Shenzhen, China
}

\begin{document}
\maketitle

\begin{center}
    \vspace{-0.5em}
    \href{https://hren20.github.io/ulvn-website/}{\texttt{https://hren20.github.io/ulvn-website}}
    \vspace{-0.5em}
\end{center}

\begin{abstract}

Image-goal navigation is a fundamental capability for embodied AI, yet its practical deployment is strained by strong prior assumptions. Existing methods predominantly rely on temporally ordered video streams or auxiliary sensors (e.g., depth, LiDAR) to maintain spatial consistency. These sequential and multimodal dependencies severely restrict scalability, especially when deploying robots using crowd-sourced or pre-recorded unordered image collections. When temporal priors are removed, current methods struggle with severe perceptual aliasing, noisy associations, and catastrophic mapping failures. To address this underexplored challenge, we propose Unordered Landmark Visual Navigation (ULVN), a unified RGB-only framework free from temporal and odometric priors. ULVN systematically mitigates error accumulation by integrating mapping, localization, and planning. Specifically, it constructs a robust 2D topological map directly from unstructured images via calibrated geometric verification and maximum spanning forest refinement. For closed-loop execution, ULVN abandons sequential heuristics, utilizing a graph-based belief propagation filter with entropy-adaptive fusion for global localization and dynamic subgoal planning. Extensive experiments in simulation and real-world deployments demonstrate that ULVN significantly outperforms state-of-the-art methods.
\vspace{1pt}
\end{abstract}
\vspace{1pt}

\section{Introduction}
\label{sec:introduction}
\begin{figure}[t]
\centering
  \begin{minipage}[b]{0.48\linewidth}
    \centering
    \includegraphics[width=\linewidth]{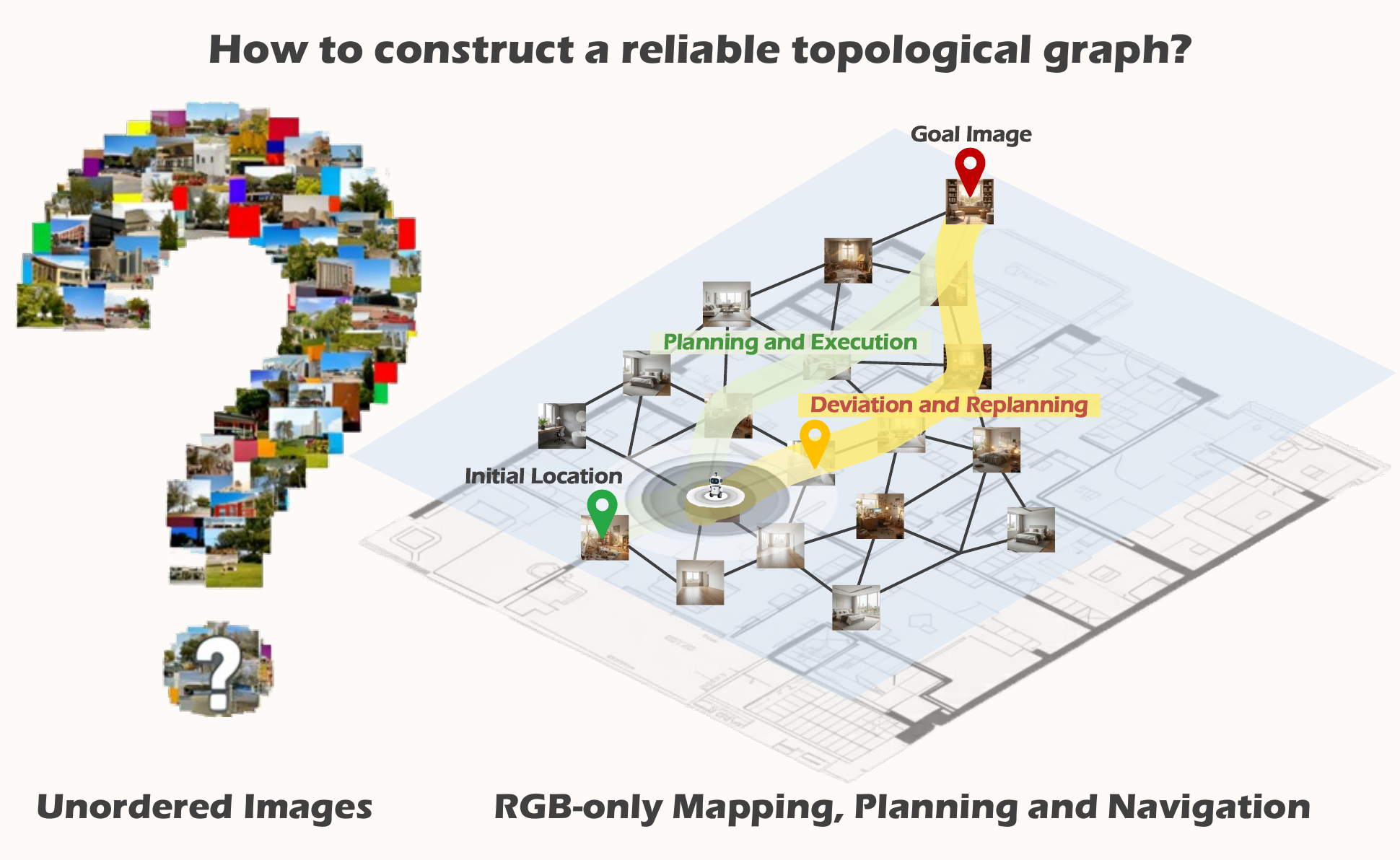}
    \vspace{-15pt}  
    \label{fig:teaser}
  \end{minipage}\hfill
  \begin{minipage}[b]{0.48\linewidth}
    \centering
    \includegraphics[width=\linewidth]{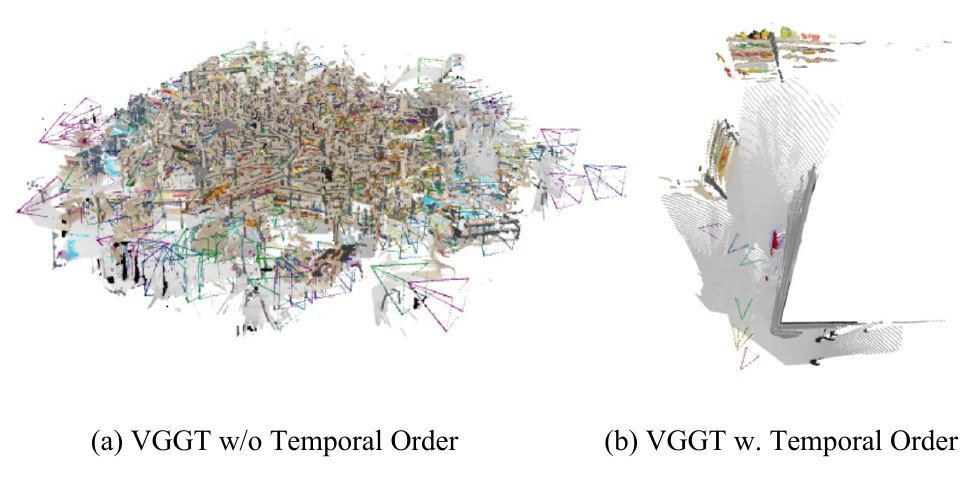}
    \vspace{-15pt}
    \label{fig:toy}
  \end{minipage}
  
  \vspace{5pt} 
  \caption{From unordered images to navigation. Our system, ULVN, automatically constructs a topological representation purely from an unstructured set of RGB images. This enables closed-loop planning, execution, and replanning. In contrast, foundation models such as VGGT struggle to handle unordered images.}
  \label{fig:1}
\vspace{-15pt}
\end{figure}

Visual navigation~\cite{bonin2008visual, furgale2010visual, zhang2022survey, gornet2024automated} is a fundamental capability for autonomous agents, supporting applications from motion planning~\cite{gao2023room} to domestic robotics~\cite{liu2024ok}. From the perspective of cognitive science, human navigation relies primarily on visual observations to form a topological understanding of the environment, enabling route finding without requiring precise metric coordinates or expensive active sensors such as LiDAR~\cite{savinov2018semi, shah2021ving}. Emulating this capability in embodied AI has therefore become a highly desirable objective~\cite{zhu2017target, anderson2018evaluation, shah2021ving}. Ideally, an embodied agent should operate predominantly on RGB observations within a pre-built topological map~\cite{booij2007navigation, shah2023vint, suomela2024placenav, thrun1998learning, cummins2008fab, savinov2018semi, cui2024frontier}, developing holistic spatial awareness and reliable goal-directed behavior.

In contrast to this ideal, many existing visual navigation systems implicitly rely on simplifying conditions that rarely hold in the wild. They often require temporally ordered video streams, which provide strong sequential priors for graph construction and policy learning~\cite{suomela2024placenav, zhang2025topological}. Absent this temporal structure, such as when using an unordered photo collection from real estate listings, learned transitions often fail. Without sequential priors, disambiguating similar places is challenging, making topological connections noisy and action selection unstable, which frequently causes oscillatory behavior. Furthermore, many systems depend on auxiliary sensors like depth or LiDAR to stabilize localization~\cite{hossain2024toponav, garg2024robohop}. Relying solely on appearance complicates geometric verification, and even advanced reconstruction models struggle with pose estimation from sparse, disconnected observations~\cite{thoma2019mapping, wang2025vggt, song2026unisharp}. Consequently, current success heavily depends on temporal continuity, auxiliary sensing, or both.

Eliminating these assumptions presents two progressive challenges. First, removing temporal priors forces topological mapping to infer connectivity from appearance alone. This vulnerability to perceptual aliasing and spurious edges makes planning and localization fragile, as errors accumulate without sequential updates to suppress ambiguity~\cite{booij2007navigation,cummins2008fab,milford2012seqslam,xu2021probabilistic}. Second, removing multimodal sensors alongside temporal priors compounds the difficulty. Without depth, LiDAR, or odometry, resolving scale and viewpoint ambiguities becomes significantly harder, increasing cumulative drift and complicating closed-loop correction using only RGB evidence~\cite{bonin2008visual,zhang2022survey,cadena2016past,yasuda2020autonomous}
. These challenges motivate our central question: \emph{Can we build an RGB-only navigation framework that achieves reliable closed-loop navigation using only unordered images, without relying on odometry, depth, or temporal priors?}

To address this question, we introduce ULVN, an unordered landmark navigation framework designed from a unified system optimization perspective. This framework ensures that mapping, localization, and planning are co-designed for consistency and reliability. For topological mapping, ULVN employs a visual place recognition and geometric verification scheme equipped with a one-shot, data-driven threshold calibration to enhance edge verification robustness. We further refine the graph by extracting a maximum spanning forest skeleton and reinserting strong loop closures to yield a reliable topological structure from unordered images. For localization, we propose a global belief propagation scheme that updates spatial probabilities through Bayesian estimation and entropy-based adaptive fusion, effectively mitigating localization drift under visual ambiguity. Finally, ULVN utilizes this global belief state to plan paths across the graph and dynamically replans when deviations occur. We comprehensively validate ULVN in both simulated and real-world environments. The main contributions of this work are as follows:

\begin{itemize}
    \item \textbf{A unified framework for unordered landmark visual navigation.} We consider RGB-only navigation without temporal or multimodal priors, an important yet still underexplored setting, and provide a unified framework that integrates mapping, localization, and navigation to help control error accumulation.
    \item \textbf{A more reliable topological mapping from unordered RGB images.} To obtain a high-quality topological map, ULVN adopts a two-stage scheme that calibrates VPR-based candidates retrieval and geometric verification using a one-shot threshold estimation and then refines the graph through an MSF skeleton with loop reinsertion.
    \item \textbf{Odometry-free global localization and navigation on 2D graphs.} ULVN updates a global belief state through Bayesian estimation with entropy-based adaptive fusion to limit drift under visual ambiguity, and uses this belief to plan and re-plan paths on the topological graph.
    \item \textbf{Benchmark and comprehensive experimental validation.} We publicly release the unordered landmark visual navigation dataset in simulation and real-world environments, together with data-collection pipelines, and evaluation metrics to support reproducible research. We provide ablation studies on error accumulation across these components and show that ULVN achieves stronger performance than existing methods.
\end{itemize}

\section{Related Work}
\label{sec:related}

\noindent
\textbf{Topological Map Construction.}
Early appearance-based visual navigation and SLAM used topological maps for efficiency and robustness, coupling visual place recognition (VPR) with probabilistic inference to mitigate perceptual aliasing~\cite{cummins2008fab,milford2012seqslam,zhang2021visual}. This line of work also indicated that image-only representations can serve as a scalable alternative to dense metric reconstructions~\cite{irschara2009structure,sattler2016efficient, fan2024navigation} in many settings.
A large fraction of recent RGB-only systems assumes temporally ordered capture~\cite{shah2023gnm,suomela2025synthetic, sun2024prioritized,ren2026strnet}, effectively yielding a one-dimensional chain of landmarks~\cite{suomela2024placenav,sridhar2024nomad,ren2025prior,gode2024flownav} and focusing on local planners~\cite{zhang2022survey,majumdar2022zson,roth2024viplanner,wan2025rapid,ren2026fisher}. While practical, these designs rely on sequential structure and are incompatible with the \emph{bag-of-images} setting where no temporal signal is available.
When sequences are absent, a 2D topological graph is typically built via VPR-based candidate retrieval followed by geometric verification~\cite{garg2021your,miao2024survey}. Classical pipelines retrieve neighbors with local features and verify edges via epipolar or homography tests~\cite{schonberger2016structure,schonberger2016pixelwise}, but are sensitive to hyperparameters, such as RANSAC inlier thresholds, which trade off edge precision and recall, ultimately affecting connectivity. Learned global descriptors improve retrieval invariance (viewpoint, illumination, season), yet descriptor similarity does not imply traversability; verification remains essential. Recent robust estimators~\cite{piedade2023bansac,wei2023generalized,barath2024stereoglue,shi2024ransac} mitigate, but do not remove, the reliance on thresholds and upper bounds across environments~\cite{edstedt2025less}.

\noindent
\textbf{Localization and Planning on Topological Maps.}
Given a map, the agent must localize and plan a path. Many planners assume precise global information (e.g., depth or metric maps)~\cite{blochliger2018topomap,muravyev2025prism}, which is often unavailable in RGB-only scenarios. A popular alternative is to estimate a ``temporal distance'' between the current and goal views~\cite{savinov2018semi,shah2021ving,ma2022vip,shah2023vint,myers2024learning,jiang2025episodic}. However, this proxy may correlate poorly with geodesic distance~\cite{neubert2019neurologically,berton2023jist,li2025casevpr}, often presumes near-constant velocity~\cite{zhao2024learning,kamila2020measuring,montano2024g}, and is expensive to evaluate over many candidates~\cite{suomela2024placenav}, while requiring robot-specific trajectory data for training~\cite{suomela2024placenav,shah2021ving,zeng2025navidiffusor}. Recent work reframes subgoal selection as VPR-based localization~\cite{suomela2024placenav,claxton2024improving}, improving efficiency and robustness, yet many models implicitly retain 1D-topology assumptions~\cite{hossain2024toponav,suomela2024placenav}, which fail on general 2D graphs with intersections, branches, and loops. In contrast to such sequential assumptions, our approach localizes on 2D graphs without odometry by using a topological transition model defined by graph adjacency.

\section{Methodology}
\label{sec:method}
To achieve image-goal navigation strictly from unordered RGB images, our framework, ULVN, adopts a purely topological approach, abandoning error-prone metric reconstruction and temporal heuristics. Assuming the provided image library contains sufficient visual overlap, ULVN systematically mitigates error accumulation through three tightly integrated modules: robust topological graph construction (RAVEL, Sec.~\ref{sec:ravel}), global state estimation (BPL, Sec.~\ref{sec:localization}), and closed-loop visual planning (BASS, Sec.~\ref{sec:bass}).

\noindent\textbf{Problem Definition and Notation.}
Let $\mathcal{L}=\{I_i\}_{i=1}^{N}$ be an unordered set of collected images, $I_g$ be the goal image, and $I_t$ be the live observation at time $t$, with no odometry or temporal priors available. A global image encoder $f(\cdot)$ produces L2-normalized embeddings $\mathbf{z}_i^r=f(I_i)$, $\mathbf{z}_g=f(I_g)$, and $\mathbf{z}_t=f(I_t)$. ULVN constructs a directed, weighted graph:
\begin{equation}
\mathcal{G} = (\mathcal{V}, \mathcal{E}, \mathbf{W}), \quad \mathcal{V} \subseteq \{1,\dots,N\}, \quad \mathbf{W} = \{W_{ij}\}_{(i,j)\in\mathcal{E}}
\end{equation}
Because metric distance is unavailable, we use visual overlap as a proxy for spatial proximity. Each edge $(i, j)\in \mathcal{E}$ has an integer weight $W_{ij} \in \mathbb{Z}_{>0}$ representing the number of verified local feature inliers.

\begin{figure*}[t]
    \centering

    \includegraphics[width=\linewidth]{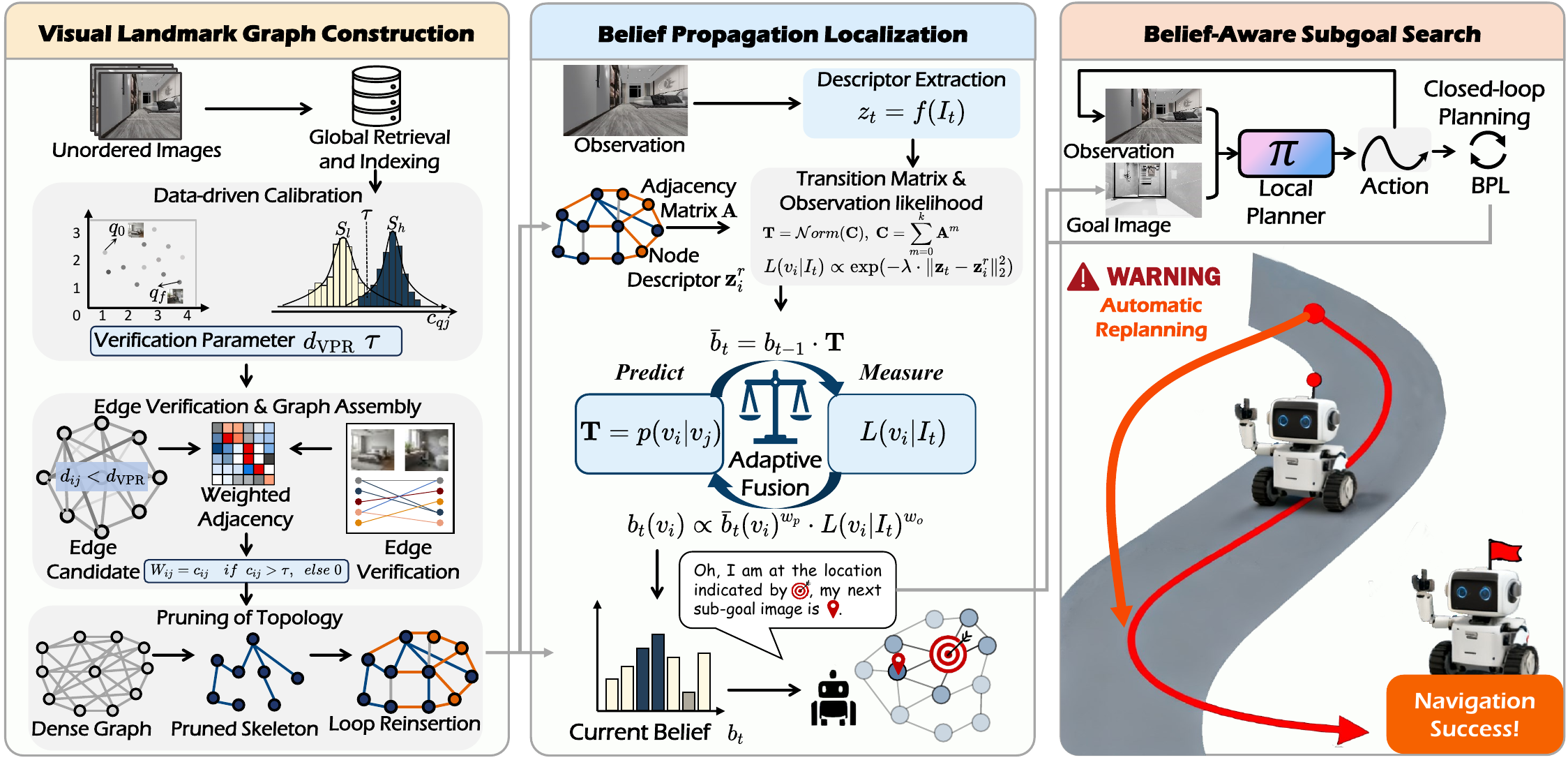}
    \vspace{-15pt}
    \caption{\textbf{Overview of Unordered Landmark Visual Navigation}. (left) build a calibrated, pruned topology from feature retrieval over unordered images. (middle) embed the observation and update node belief with observation likelihoods and transition \(\mathbf{T}\) via adaptive fusion. (right) pick the next landmark from the belief and plan in closed loop; belief shifts trigger automatic replanning to the goal.}
    \label{fig:main}
    \vspace{-15pt}
\end{figure*}

\subsection{Visual Landmark Graph Construction}
\label{sec:ravel}

We introduce RAVEL (Robust Augmentation and VErification of Landmarks), a pipeline that transforms $\mathcal{L}$ into a weighted, directed visual graph. Unordered collections naturally produce dense, redundant candidate pairs that can cause memory bloat and degrade downstream planning. RAVEL mitigates this by applying global retrieval for fast recall, followed by geometric verification as a high-precision bottleneck, and ultimately pruning the graph into a compact, structurally meaningful skeleton.

\textbf{Global Retrieval and Indexing.}
To avoid exhaustive pairwise matching, we first extract an L2-normalized global descriptor $\mathbf{z}_i^r = f(I_i)$ for each image using an image encoder $f(\cdot)$. While $f(\cdot)$ can be a general-purpose backbone (e.g., ResNet, DINOv2), our experiments show that representations trained explicitly for Visual Place Recognition (VPR) yield the most robust retrieval candidates. All global descriptors are indexed using FAISS~\cite{johnson2019billion, douze2025faiss}.

To automatically calibrate verification thresholds, we select the first image $q_0$ and retrieve its farthest valid neighbor $q_f = \arg\max_{j \in \mathcal{N}_{\text{FAISS}}(q_0)} \|\mathbf{z}_{q_0} - \mathbf{z}_j\|_2^2$, explicitly excluding invalid index outputs.
For each anchor $q \in \{q_0, q_f\}$, we perform full feature matching against all retrieved images, yielding an inlier count $c_{qj}$ and anchor descriptor distance $d_{qj}$ for each pair. Applying $k$ mean clustering to pooled inlier counts yields a cluster of highest-confidence $S_h$ and a cluster of second-highest-confidence $S_l$. The inlier threshold $\tau$ is defined as:
\begin{equation} \label{eq:tau}
\tau = \frac{1}{2}\big(\min(S_h) + \max(S_l)\big).
\end{equation}
The global distance threshold is derived from $S_h$: $d_{\text{VPR}} = \max_{(q,j) \in S_h} \|\mathbf{z}_q - \mathbf{z}_j\|_2$.

\textbf{Local Geometric Verification and Graph Pruning.}
For every candidate pair where the global descriptor distance $s_{ij} \le d_{\text{VPR}}$, we perform local geometric verification using a deep matcher (e.g., LightGlue~\cite{lindenberger2023lightglue}) to obtain the precise inlier match count $c_{ij}$. The weighted adjacency matrix is updated as:
\begin{equation} \label{eq:weight}
W_{ij}=W_{ji}= \begin{cases}
c_{ij},\quad & c_{ij} > \tau\\
0,\quad & \text{otherwise}.
\end{cases}
\end{equation}

Raw graphs built from unordered collections are inherently dense and contain weak, redundant edges that not only cause memory bloat but also degrade the subsequent belief propagation by diffusing probability mass into noisy pathways. To extract a compact, reliable skeleton, we compute the Maximum Spanning Forest (MSF) $\mathcal{F} = (\mathcal{V}, \mathcal{E}_{\text{MSF}})$. By maximizing the total edge weight $\sum_{(i,j) \in \mathcal{E}_{\text{MSF}}} W_{ij}$, the MSF ensures the navigational backbone consists strictly of the strongest geometric transitions, minimizing local execution failures.

While the MSF effectively eliminates visual aliases, it destroys cyclic structures essential for alternate routing and topological error correction. To recover these, we introduce data-driven \emph{Strong-Loop Reinsertion}. Instead of brittle heuristics, we cluster all edge weights in $\mathcal{G}$ into $k=10$ clusters via $k$-means. We define a dynamic threshold $\tau_{\text{add}}$ as the minimum weight within the two highest-centroid clusters to ensure adaptability to different scenes. Reinserting excluded edges where $W_{ij} > \tau_{\text{add}}$ yields the final robust graph $\mathcal{G}_{\text{pruned}}$, seamlessly balancing structural reliability with critical topological redundancy.

\subsection{Belief Propagation Localization}
\label{sec:localization}

Building upon the foundations of classical Bayesian filtering~\cite{thrun2002probabilistic}, we extend the 1D sequential localization filters utilized in prior works~\cite{xu2021probabilistic, suomela2024placenav} to operate on arbitrary 2D spatial graphs with complex branching and loops. We introduce a Belief Propagation Localization (BPL) filter that maintains a belief distribution $b_t$ over all nodes $v_i \in \mathcal{V}$ via a recursive prediction-correction cycle.

\textbf{Topological Prediction Step.}
Let $\mathbf{A}$ be the binary adjacency matrix derived from $\mathcal{G}_{\text{pruned}}$. To model the robot's possible motion over multiple steps without odometry, we define a cumulative reachability matrix aggregating paths up to length $K=3$:
\begin{equation} \label{eq:cumulative}
\mathbf{C} = \sum_{m=0}^{K} \mathbf{A}^m,
\end{equation}
where $\mathbf{A}^0 = \mathbf{I}$. Row-normalizing $\mathbf{C}$ yields a stochastic transition matrix $\mathbf{T}_{ij} = \mathbf{C}_{ij} / \sum_{j'}\mathbf{C}_{ij'}$. The predicted belief is $\bar{b}_t = b_{t-1} \cdot \mathbf{T}$.

\textbf{Adaptive Observation Fusion.}
The observation likelihood is defined as $L(v_i | I_t) = \exp(-\lambda \cdot \|\mathbf{z}_t - \mathbf{z}_i^r\|_2^2)$ with $\lambda = 5$. We introduce an entropy-adaptive fusion mechanism. We quantify uncertainty using the normalized Shannon entropy of the prior belief:
\begin{equation} \label{eq:entropy}
H_n(b_{t-1}) = -\frac{1}{\log|\mathcal{V}|} \sum_{v_i \in \mathcal{V}} b_{t-1}(v_i) \log b_{t-1}(v_i).
\end{equation}

To dynamically balance prediction and observation, we set the fusion weights directly from the entropy such that $w_p = 1 - H_n(b_{t-1})$ and $w_o = H_n(b_{t-1})$, ensuring they strictly sum to $1$. The posterior belief is computed via a normalized weighted geometric mean:
\begin{equation} \label{eq:posterior}
b_t(v_i) \propto \bar{b}_t(v_i)^{w_p} \cdot L(v_i | I_t)^{w_o}.
\end{equation}

The estimated location is the Maximum A Posteriori (MAP) node: \\
$\hat{v}_t = \arg\max_{v_i} b_t(v_i)$.

\subsection{Belief-Aware Subgoal Search for Navigation}
\label{sec:bass}

With the graph constructed and the localization filter initialized, Belief-Aware Subgoal Search (BASS) plans and executes paths entirely from images.

\textbf{Task Initialization.}
A navigation task begins with a goal image $I_g$. Before planning, we must anchor the start and goal states on the graph $\mathcal{G}$. We establish the start node $v_s$ using the maximum-a-posteriori estimate of the initial global belief $b_0$: $v_s = \arg\max_{v_i \in \mathcal{V}} b_0(v_i)$. The goal node $v_g$ maximizes descriptor similarity: $v_g = \arg\max_{v_i \in \mathcal{V}} (\mathbf{z}_g^\top \mathbf{z}_i^r)$.

\textbf{Topological Planning.}
We treat $\mathcal{G}$ as a weighted graph where edge weights $W_{ij}$ reflect visual overlap. Because a visual navigation path is only as robust as its weakest visual link, we define the confidence of a candidate path $\mathcal{P}$ on the graph $\mathcal{G}$ from $v_s$ to $v_g$ as its minimum edge weight:
\begin{equation} \label{eq:path_conf}
\mathrm{conf}(\mathcal{P}) = \min_{(v_i, v_j) \in \mathcal{P}} W_{ij}.
\end{equation}
BASS extracts the widest path $\mathcal{P}^* = \arg\max_{\mathcal{P}} \mathrm{conf}(\mathcal{P})$ with the Dijkstra's algorithm . This max-min objective guarantees that the chosen sequence of visual subgoals $(v_s, v_1, \dots, v_g)$ has the highest probability of successful image-to-image local control.

\textbf{Closed-Loop Vision-Only Execution.}
Our framework operates strictly without odometry. The robot follows $\mathcal{P}^*$ by sequentially passing the planned subgoals to an image-based local planner (e.g., a visual foundation model like ViNT~\cite{shah2023vint} or a generic visual servoing policy). At each step, the local planner takes the observation $I_t$ and the current topological subgoal image $I_{s_n}$ and directly outputs low-level velocity commands to drive the robot toward the target.

During execution, BPL continuously updates the belief $b_t$. A deviation is detected if the MAP node $v_t$ falls off the planned path $\mathcal{P}^*$ or if the graph distance from $v_t$ to the current subgoal exceeds $D_{\text{thres}} = 3$. Upon deviation, BASS dynamically replans a new widest path from $v_t$ to $v_g$. Navigation succeeds when the belief converges on the goal node with a probability exceeding $0.5$.

\section{Experiments}
\label{sec:experiment}
\begin{figure}[t]
\begin{center}
\includegraphics[width=0.6\linewidth]{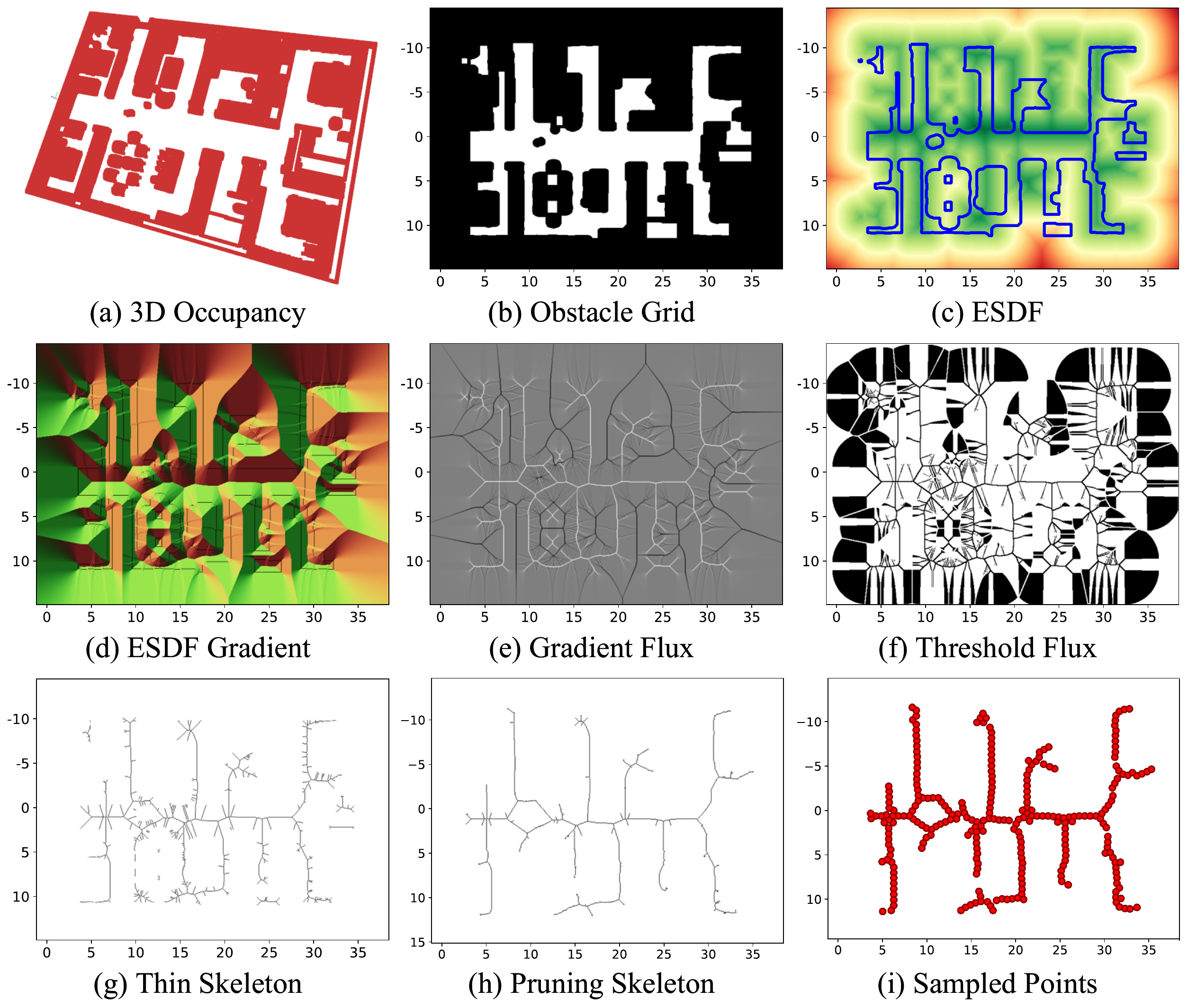}
\end{center}
\vspace{-10pt}
\caption{The topological skeleton extraction in data collection pipeline. It starts with (a) a 3D occupancy map, which is projected into (b) a 2D binary obstacle grid. (c) A Euclidean Signed Distance Function (ESDF) and (d) its gradient. (e) The ESDF gradient flux is then calculated, (f) thresholded to isolate the medial axis, (g) thinned, (h) pruned to produce the clean topological skeleton, and finally (i) indicates the node of acquired images.}
\label{fig:datacollection}
\vspace{-10pt}
\end{figure}

\subsection{Experimental Settings}
\label{sec:expset}
\textbf{Evaluation Scenes.} We evaluate our method, ULVN, within the NVIDIA Isaac Sim simulator using the GRScenes dataset~\cite{wang2024grutopia}, a high-fidelity benchmark for general-purpose robotics. We utilize 10 distinct scenes, equally split between \textit{home} and \textit{commercial} environments. All scenes feature realistic physics and materials. We also employ the CARLA~\cite{dosovitskiy2017carla} simulator to specifically isolate and validate the RAVEL and BPL modules under varying perceptual conditions.

\noindent
\textbf{Tasks and Metrics.}
We evaluate ULVN across three core capabilities:
1) \textit{Topomap Construction:} We report Precision ($P$), Recall ($R$), F1-score ($F1$), and Accuracy ($Acc.$) of the generated graph edges evaluated against the ground-truth spatial connectivity (derived via Sec.~\ref{sec:datacollection}). We also analyze the Pearson correlation ($r$) between global descriptor similarity and true geometric inliers to justify our retrieval design.
2) \textit{Localization:} We report the localization Accuracy (success rate of the MAP estimate correctly identifying the nearest topological node) when localizing a continuous stream of noisy images against the pre-built, non-sequential graph.
3) \textit{Navigation:} We adopt standard Habitat ImageNav metrics~\cite{savva2019habitat}: Success Rate (SR), Success weighted by Path Length (SPL), and average number of collisions.

\noindent
\textbf{Baselines.}
We compare against strong state-of-the-art representations and systems.
For \textit{Topomap Construction}, we evaluate ResNet-50~\cite{he2016deep}, DINOv2~\cite{oquab2023dinov2}, MegaLoc~\cite{berton2025megaloc}, ViNT~\cite{shah2023vint}, and PlaceNav~\cite{suomela2024placenav}. Because ViNT and PlaceNav traditionally rely on temporally ordered videos, we adapt them for unordered mapping by using their temporal distance predictions or VPR retrieval scores as direct edge-weight proxies.
For \textit{Localization}, we compare against single-frame VPR (MegaLoc, PlaceNav), temporal distance (ViNT), and sequence smoothing (JIST~\cite{berton2023jist}).
For \textit{Navigation}, we evaluate the integration of our framework with visual local planners (ViNT, NoMaD) and compare with other SOTA methods.

\noindent
\textbf{Implementation Details.}
Global descriptors are extracted using MegaLoc~\cite{berton2025megaloc} with L2-normalization, and geometric verification is performed via LightGlue~\cite{lindenberger2023lightglue}. During RAVEL calibration, if the high-confidence cluster $S_h$ is valid ($k=2$), $\tau$ and $d_{\mathrm{VPR}}$ are dynamically set from its statistics; otherwise, we default to $\tau_{\text{default}}=15$ and $d_{\text{VPR,default}}=1.7$. The MSF is computed using Kruskal's algorithm, with strong-loop reinsertion bounded by $\tau_{\text{add}}=1.5\tau$. For BPL, the observation likelihood scaling factor is $\lambda = 10$.

\begin{figure*}[t]
\begin{center}
\includegraphics[width=\linewidth]{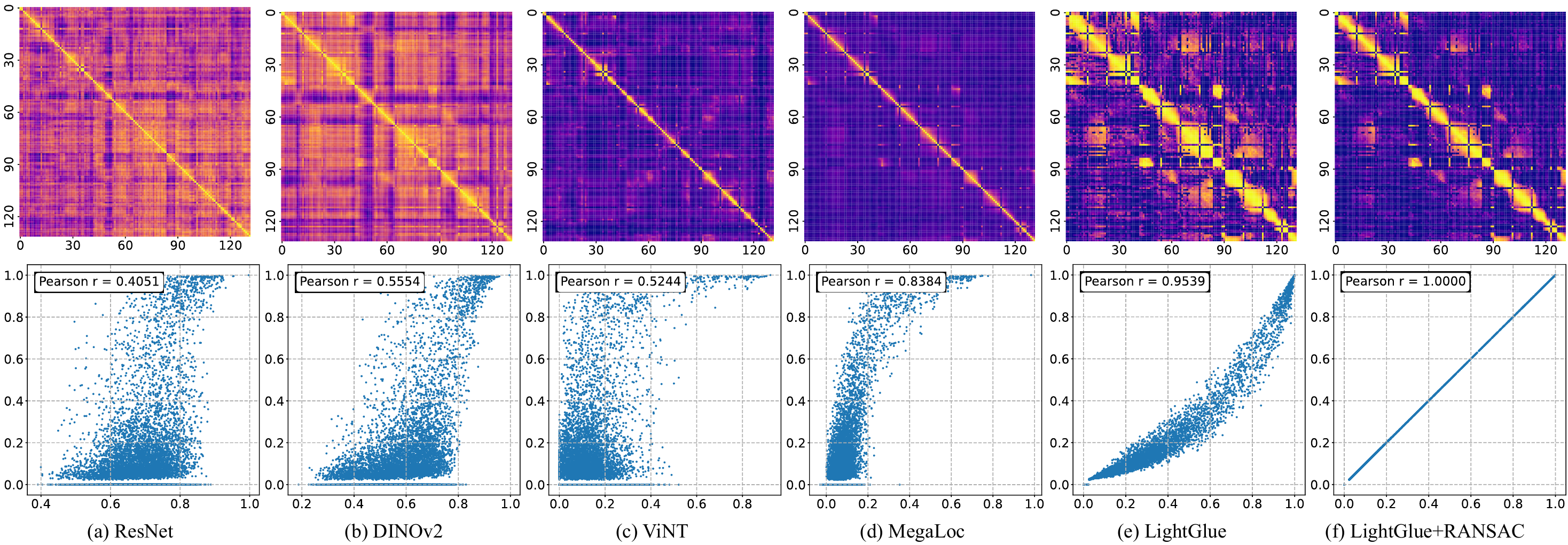}
\end{center}
\vspace{-15pt}
\caption{\textbf{Global Descriptors as Efficient Proxies for Geometric Verification}. We compare pairwise similarity scores from various global descriptors: (a) ResNet, (b) DINOv2, (c) ViNT (temporal distance), and (d) MegaLoc (VPR-trained), against a ground-truth geometric consistency score from (f) LightGlue+RANSAC. Top row: Pairwise affinity matrices for an unordered image set. Bottom row: Correlation plots (with Pearson's $r$) for each descriptor's score (Y-axis) versus the ground truth (X-axis). The VPR-trained MegaLoc (d) demonstrates a strong linear correlation, validating it as an effective and efficient proxy for geometric verification.}

\label{fig:ravel}
\vspace{-10pt}
\end{figure*}

\begin{table}[t]
\centering
\caption{Comparison of different retrieval baselines and our method on GRScenes.}
\vspace{-10pt}
\setlength{\tabcolsep}{2pt}
\renewcommand{\arraystretch}{0.9}
\begin{tabular}{lcccc}
\toprule
\textbf{Method} & \textbf{P} & \textbf{R} & \textbf{F1} & \textbf{Acc.} \\
\midrule
\textbf{VGGT}~\cite{wang2025vggt} & 0.1599 & 0.1659 & 0.1629 & 0.9931 \\
\textbf{PlaceNav top 2}~\cite{suomela2024placenav} & 0.5262 & 0.7113 & 0.6043 & 0.9948 \\
\textbf{PlaceNav top 5}~\cite{suomela2024placenav} & 0.2699 & 0.7655 & 0.3731 & 0.9853 \\
\textbf{ViNT top 2}~\cite{shah2023vint}     & 0.5201 & 0.6775 & 0.5815 & 0.9946 \\
\textbf{ViNT top 5}~\cite{shah2023vint}     & 0.2660 & \textbf{0.8246} & 0.3806 & 0.9816 \\
\textbf{RAVEL (Ours)} & \textbf{0.7104} & 0.7656 & \textbf{0.7365} & \textbf{0.9970} \\
\bottomrule
\end{tabular}
\vspace{-15pt}
\label{tab:ravel}
\end{table}

\begin{table}[t]
\centering
\caption{Quantitative evaluation of topological graph construction on GRScenes. Ablations compare our RAVEL with a Top-$k$ ANN retrieval baseline and the variant ablation of our method.}
\vspace{-10pt}
\setlength{\tabcolsep}{1pt}

\begin{tabular}{lcccc}
\toprule
\textbf{Method} & \textbf{P} & \textbf{R} & \textbf{F1} & \textbf{Acc.} \\
\midrule
\textbf{Top-$k$ ANN} & 0.1245 & \textbf{0.9121} & 0.2156 & 0.9584 \\
\textbf{RAVEL w/o (MSF, AVP $\tau$)} & 0.3676 & 0.7177 & 0.4496 & 0.9898 \\
\textbf{RAVEL w/o MSF} & 0.6910 & 0.4220 & 0.5157 & 0.9956 \\
\textbf{RAVEL (Ours)} & \textbf{0.7104} & 0.7656 & \textbf{0.7365} & \textbf{0.9970} \\
\bottomrule
\end{tabular}
\label{tab:ravel_ablation}

\end{table}

\subsection{Data Collection and Ground Truth Pipeline}
\label{sec:datacollection}

To rigorously evaluate Topomap metrics, ground-truth spatial connectivity is required. As shown in Fig.~\ref{fig:datacollection}, we developed an automated in-simulation pipeline that exports 3D scene occupancy and converts it into a 2D traversability grid based on the robot's dimensions. Following~\cite{li2025learning}, we use the Euclidean Signed Distance Field (ESDF)~\cite{noel2023skeleton} to extract the topological skeleton of the traversable area. By applying non-maximum suppression along this skeleton, we sample sparse and spatially uniform nodes. The underlying skeletal structure provides the absolute ground-truth connectivity matrix used to evaluate the precision and recall of RAVEL's generated graphs. Further details are provided in the Appendix.

\begin{table}[t]
\centering

\caption{Robustness of RAVEL}
\vspace{-10pt}

\label{tab:ravel_noise}
\resizebox{\linewidth}{!}{%
\begin{tabular}{lcccc}
\hline
\textbf{Method} & \textbf{Precision} & \textbf{Recall} & \textbf{F1-Score} & \textbf{Accuracy}\\
\hline
\textbf{RAVEL+noise} &
0.6703 (-5.64\%) & 0.7072 (-7.63\%) & \textbf{0.6882} (\textbf{-6.56\%}) & \textbf{0.9963} (\textbf{-0.07\%})\\
\textbf{CosPlace top 2+nois}e &
0.4592 (-12.73\%) & 0.6615 (-7.00\%) & 0.5420 (-10.29\%) & 0.9936 (-0.12\%) \\
\textbf{CosPlace top 5+noise} &
0.2192 (-18.78\%) & 0.7739 (+1.10\%) & 0.3416 (-8.44\%) & 0.9826 (-0.27\%)\\
\textbf{ViNT top 2+noise}    & 0.4197 (-19.30\%) & 0.6064 (-10.49\%) & 0.4961 (-15.70\%) & 0.9913 (-0.33\%) \\
\textbf{ViNT top 5+noise}    & 0.2065 (-22.36\%) & 0.7976 (-3.27\%)  & 0.3281 (-18.43\%) & 0.9769 (-0.48\%) \\

\hline
\end{tabular}%
}
\vspace{-10pt}
\end{table}

\subsection{Evaluation of RAVEL}

\textbf{Global Descriptors as Proxies for Geometric Verification.}
We first assess the viability of using global descriptors as a fast-recall proxy for costly geometric verification. Using CARLA data, we analyze the Pearson correlation ($r$) between descriptor similarity and the ground-truth geometric score (LightGlue + RANSAC inliers). As shown in Figure~\ref{fig:ravel}, general-purpose features like ResNet (a) and DINOv2 (b) exhibit weak, non-linear correlations, leading to severe perceptual aliasing and false-positive retrievals. Similarly, ViNT's temporal distance (c) fails to reliably map to geometric overlap in unordered settings. Conversely, the VPR-trained MegaLoc (d) demonstrates a strong, tightly-clustered linear correlation with geometric reality. This confirms that VPR is strictly necessary for the initial retrieval stage to filter candidates effectively without discarding valid topological neighbors.

\noindent
\textbf{Mapping Performance Evaluation.}
Table~\ref{tab:ravel} and Table~\ref{tab:ravel_ablation} detail the graph construction performance across 10 GRScenes. Retrieval-only baselines (CosPlace, ViNT) highlight a fundamental tension: increasing the Top-$k$ threshold boosts recall but decimates precision, yielding poor F1 scores. This indicates that raw appearance similarity cannot substitute for structural verification.

Our ablation (Table~\ref{tab:ravel_ablation}) reveals the mechanics of our solution. The adaptive-threshold variant (\textit{Ours w/o MSF}) achieves high precision by pruning weak visual matches, but suffers from low recall because purely local edge rejection fragments the graph. Applying the MSF resolves this. By shifting the objective from isolated edge evaluation to global structural coherence, the MSF reconstructs a unified skeleton. This allows RAVEL to aggressively filter false positives while maintaining high recall, ultimately achieving the highest F1-score and demonstrating robust generalization across diverse scenes.

\noindent
\textbf{Robustness to Visual Perturbations.}
To evaluate the robustness of our graph construction against real-world sensor imperfections, we introduce a \textit{+noise} setting (Table~\ref{tab:ravel_noise}). We apply a suite of random visual perturbations to the images prior to mapping. This includes lighting variations (random brightness scaling $\in [0.75, 1.25]$, RGB color shifting $\pm 5\%$, contrast adjustments $\in [0.85, 1.15]$, and low-light Gaussian noise with $\sigma \in [3, 10]$ triggered when brightness drops below $0.9$) and directional motion blur (random kernel lengths $\in \{3, 5, 7, 9, 11\}$ at angles $\in \{0^\circ, 45^\circ, 90^\circ, 135^\circ\}$).

As shown in Table~\ref{tab:ravel_noise}, purely retrieval-based methods with VPR (CosPlace) and temporal distance (ViNT) are highly sensitive to these pixel-level perturbations, suffering massive drops in precision and F1-score. This confirms that appearance-based distances alone degrade rapidly under environmental noise. In contrast, RAVEL demonstrates remarkable resilience, experiencing only a $6.56\%$ drop in F1-score. While the noise inevitably reduces the raw number of local feature inliers, RAVEL's geometric verification and the MSF's global structural constraints successfully prevent the corrupted images from forming spurious edges, thereby preserving a high accuracy and structural integrity.

\subsection{Evaluation of BPL}
\begin{table}[t]
\centering

\caption{Quantitative comparison of global localization under Global and Difficult Path scenarios.}
\vspace{-10pt}
\setlength{\tabcolsep}{1pt}
\renewcommand{\arraystretch}{0.9}
\begin{tabular}{clcccc}
\toprule
\textbf{Scenario} & \textbf{Metric} & \textbf{Ours} & \textbf{MegaLoc~\cite{berton2025megaloc}} & \textbf{ViNT~\cite{shah2023vint}} & \textbf{JIST}~\cite{berton2023jist} \\
\midrule
\multirow{4}{*}{\textbf{All}}
 & Nodes & 976 & 976 & 976 & 976\\
 & Success & \textbf{932} & 889 & 845 & 829\\
 & Fail & \textbf{44}  & 87  & 131 & 147\\
 & Acc.(\%) & \textbf{95.49} & 91.09 & 86.58 & 84.94\\
\midrule
\multirowcell{4}{\textbf{Difficult}\\\textbf{Path}}
 & Nodes & 383 & 383 & 383 & 383\\
 & Success & \textbf{360} & 341 & 270 & 315\\
 & Fail & \textbf{23}  & 42  & 113 & 68\\
 & Acc.(\%) & \textbf{93.99} & 89.03 & 70.50 & 82.25\\
\bottomrule
\label{tab:bpl}
\end{tabular}
\vspace{-10pt}
\end{table}

\begin{table*}[t]
  \centering
  \caption{Comparison of localization accuracy of different methods under different conditions. Results are reported as mean $\pm$ standard deviation across four datasets.}
  \vspace{-10pt}
  \label{tab:localization_noise_average}
  \setlength{\tabcolsep}{4pt}
  \renewcommand{\arraystretch}{0.9}
  \begin{tabular}{lcccc}
    \toprule
    \multirow{2}{*}{\textbf{Condition}} & \multicolumn{4}{c}{\textbf{Localization Accuracy}} \\
    \cmidrule(lr){2-5}
    & \textbf{MegaLoc}~\cite{berton2025megaloc} & \textbf{JIST}~\cite{berton2023jist} & \textbf{ViNT}~\cite{shah2023vint} & \textbf{BPL (Ours)} \\
    \midrule
    Rotation      & 0.913 $\pm$ 0.104 & 0.722 $\pm$ 0.066 & 0.895 $\pm$ 0.044 & \textbf{0.966 $\pm$ 0.020} \\
    Rot + Gauss   & 0.743 $\pm$ 0.088 & 0.452 $\pm$ 0.177 & 0.885 $\pm$ 0.045 & \textbf{0.914 $\pm$ 0.038} \\
    Rot + Poisson & 0.717 $\pm$ 0.114 & 0.445 $\pm$ 0.176 & 0.888 $\pm$ 0.037 & \textbf{0.896 $\pm$ 0.036} \\
    Rot + Crop    & 0.855 $\pm$ 0.151 & 0.525 $\pm$ 0.076 & 0.640 $\pm$ 0.096 & \textbf{0.945 $\pm$ 0.027} \\
    \midrule
    \textbf{Average} & 0.807 $\pm$ 0.133 & 0.536 $\pm$ 0.167 & 0.827 $\pm$ 0.124 & \textbf{0.930 $\pm$ 0.040} \\
    \bottomrule
  \end{tabular}
  \vspace{-10pt}
\end{table*}

\begin{table}[t]
    \centering
    \caption{Ablation Study for Belief Propagation Localization}
    \vspace{-10pt}
    \setlength{\tabcolsep}{1pt}
    \renewcommand{\arraystretch}{0.9}
    \begin{tabular}{ccccccc}
        \toprule
        \textbf{$k$} & 1 & 2 & 3 & 4 & 5 & 1\\
        \midrule
        \textbf{Adaptive Fusion} & \ding{51} & \ding{51} & \ding{51} & \ding{51} & \ding{51} & \ding{55}\\
        \midrule
        \textbf{Acc.} (\%) & 95.49 & 94.47 & 93.65 & 92.42 & 92.73 & 90.57\\
        \bottomrule
    \end{tabular}
    \label{tab:localizationablation}
    \vspace{-10pt}
\end{table}

To evaluate localization, we simulate robot motion by collecting continuous image streams along random trajectories in GRScenes. Crucially, these test images are captured at novel poses distinct from the reference nodes in the topological graph. We introduce a \textit{Difficult Path} subset, characterized by long routes, large-angle turns, and low start-to-goal visual overlap.

\noindent
\textbf{Comparison with Other Paradigms.}
As shown in Table~\ref{tab:bpl}, BPL delivers the highest accuracy. While baselines degrade noticeably on complex routes, BPL's performance remains stable. 1D temporal-distance methods like ViNT fail significantly on challenging paths with intersections or loops. BPL's robustness stems from its graph-derived transition matrix, which seamlessly diffuses belief over multi-hop neighborhoods, allowing probability mass to branch at junctions and reconverge during loop closures.

\noindent
\textbf{Ablation for BPL.}
Table~\ref{tab:localizationablation} analyzes BPL's internal mechanisms. Modifying the propagation depth ($k$) shows that performance is relatively stable, peaking around 2--3 hops; excessive diffusion over-smooths the belief and slightly increases errors. More importantly, disabling the entropy-adaptive fusion causes a severe accuracy drop from $0.9549$ to $0.9057$. This proves that while structural diffusion is helpful, dynamically weighting the prediction and observation based on current uncertainty is the critical factor for resolving visual ambiguity.

\noindent
\textbf{Robustness to Perceptual Degradation.}
To evaluate the resilience of BPL against real-world sensor noise and viewpoint shifts during live execution, we test localization accuracy under severe image perturbations (Table~\ref{tab:localization_noise_average}). We use continuous trajectories from each of four diverse datasets: RECON~\cite{shah2021rapid}, SCAND~\cite{karnan2022scand}, GoStanford~\cite{hirose2019deep}, and SACSoN~\cite{hirose2023sacson}. The live query images are subjected to four degradation conditions: pure angular viewpoint shift (Rotation), and rotation compounded with Gaussian noise, Poisson noise, or random structural cropping.

As shown in Table~\ref{tab:localization_noise_average}, BPL achieves the highest and most stable localization accuracy (averaging $0.930 \pm 0.040$) across all degradation conditions. While pure VPR (MegaLoc), sequence smoothing (JIST), and temporal models (ViNT) exhibit severe vulnerability to compounded pixel noise or structural occlusions, BPL consistently maintains an accuracy above $0.89$. This resilience validates our entropy-adaptive fusion design: when severe perceptual degradation flattens the observation likelihood, the filter dynamically shifts its reliance to the topological prediction, leveraging the multi-hop spatial prior to coast through temporary visual disruptions without losing track.

\subsection{Evaluation of BASS}

\begin{figure}[t]
\centering
  \begin{minipage}{0.48\linewidth}
    \centering
    \includegraphics[width=\linewidth]{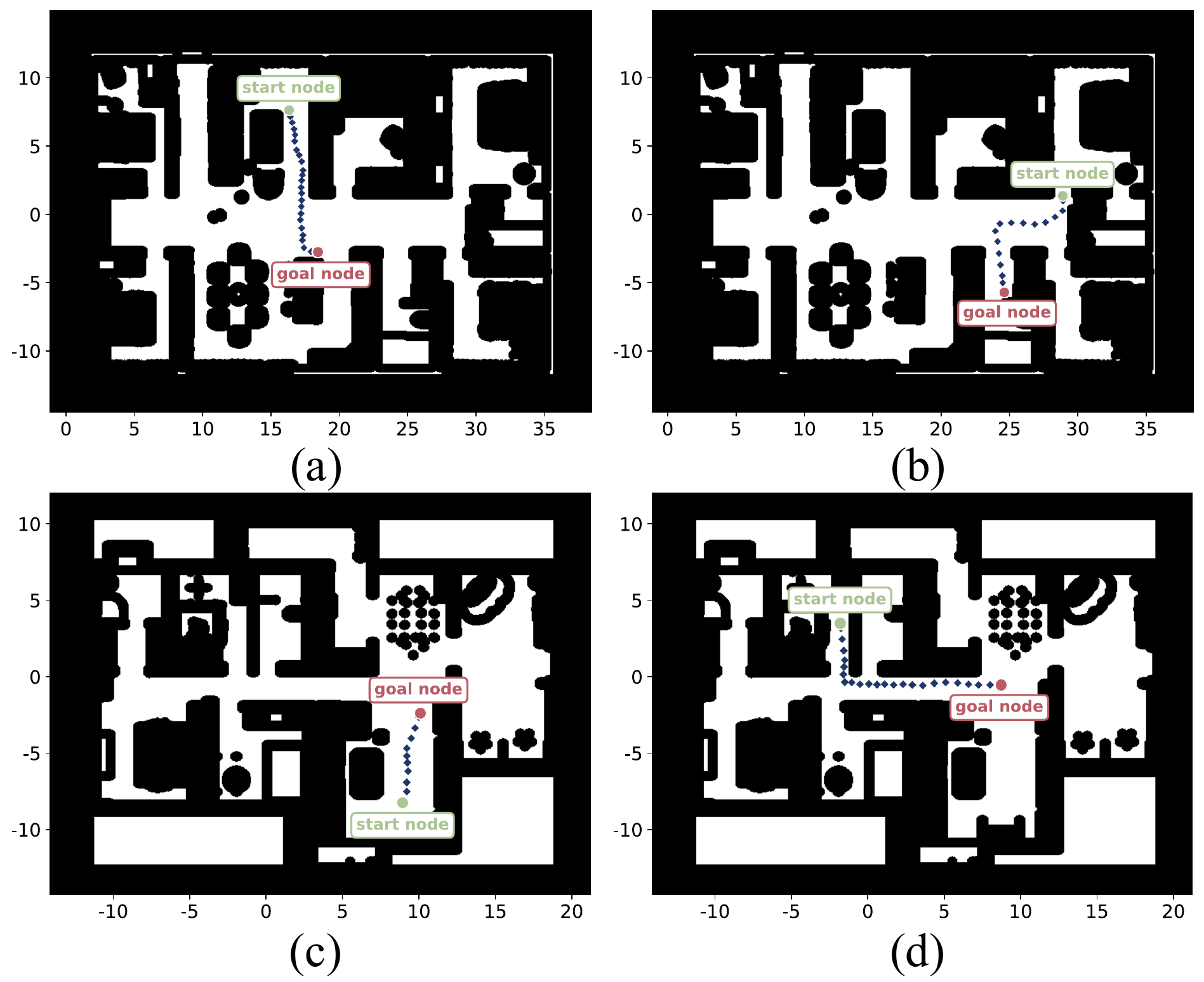}
    \vspace{-15pt}
    \caption{BASS navigation in easy (a,c) and hard (b,d) tasks. Green: start; red: goal; blue: executed trajectory.}
    \label{fig:navigation}
  \end{minipage}\hfill
  \begin{minipage}{0.48\linewidth}
    \centering
    \includegraphics[width=\linewidth]{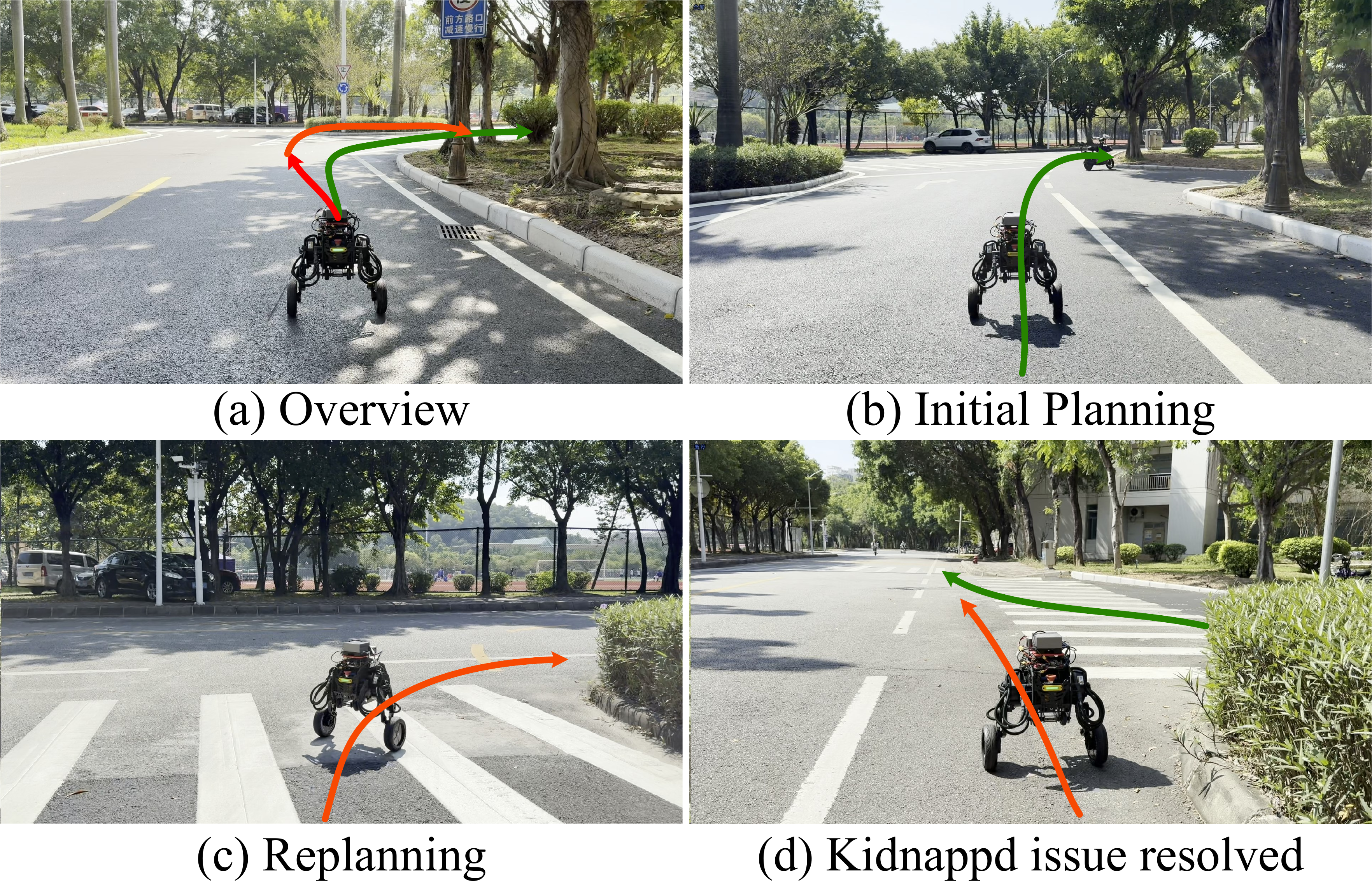}
    \vspace{-15pt}
    \caption{Real-world case. (a) Overview. (b) Initial plan. (c) Replanning after deviation. (d) Kidnapped-robot case resolved via relocalization and replanning. Green: planned path; red: observed deviation; orange: replanned path.}
    \label{fig:realworld}
  \end{minipage}
\vspace{-10pt}
\end{figure}
\begin{table}[t]
    \centering
    \caption{Comparison of Different Algorithms on GRScenes. "\textbf{-A}" means retraining with only the action head loss function. "\textbf{w. $d_{temp}$}" denotes that localization with temporal distance instead of BPL.}
    \vspace{-10pt}
    \label{tab:comparison}
    \setlength{\tabcolsep}{1pt}
    \renewcommand{\arraystretch}{0.9}
    \begin{tabular}{lccc}
        \toprule
        \textbf{Method} & \textbf{SR} (\%) & \textbf{Avg. Colli.} & \textbf{Avg. SPL} \\
        \midrule
        \textbf{Uni-Navid}~\cite{zhang2024uni} & 32.0 & 1.96 & 0.2391 \\
        \textbf{UniGoal}~\cite{yin2025unigoal} & 61.6 & 0.88 & 0.3176 \\
        \textbf{ULVN+VINT-A}~\cite{shah2023vint} & 31.0 & 1.13 & 0.812  \\
        \textbf{ULVN+NoMaD-A}~\cite{sridhar2024nomad} & 54.3 & 0.94 & 0.7458 \\
        \textbf{ULVN+ViNT w. $d_{temp}$}~\cite{shah2023vint} & 59.6 & 0.88 & 0.7752 \\
        \textbf{ULVN+ViNT}~\cite{shah2023vint} & 68.1 & 0.76 & \textbf{0.8398} \\
        \textbf{ULVN+NoMaD}~\cite{sridhar2024nomad} & \textbf{71.9} & \textbf{0.42} & 0.7978 \\
        \bottomrule
    \end{tabular}
    \vspace{-10pt}
    \label{tab:navigation}
\end{table}

\begin{figure}[htbp]
    \centering

    \includegraphics[width=\linewidth]{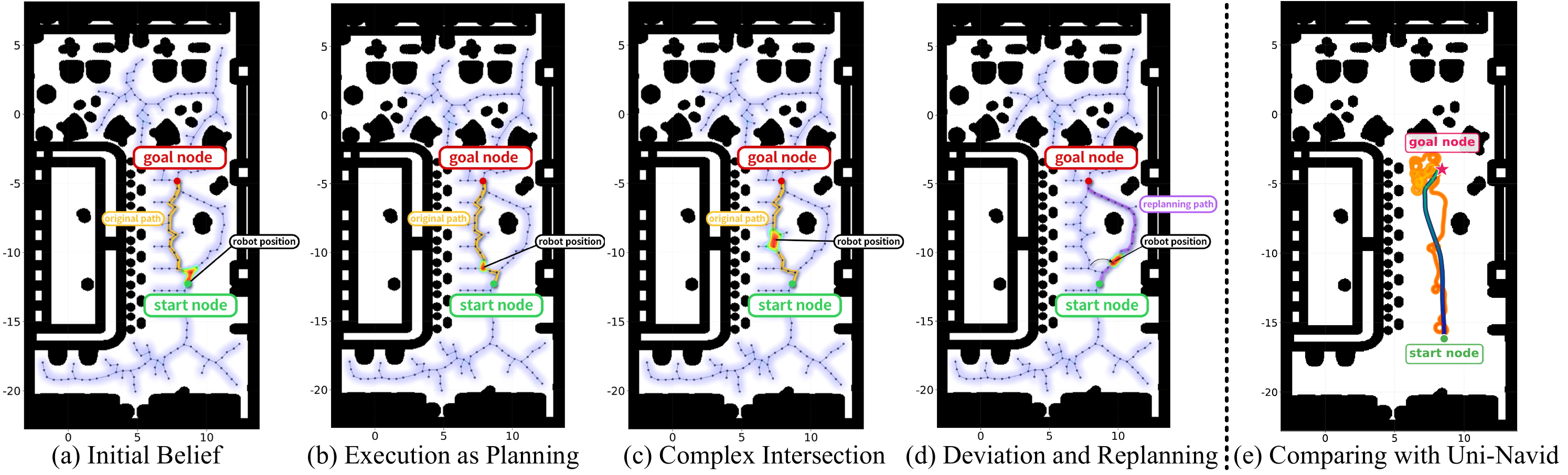}
    \vspace{-10pt}
    \caption{Visualization of replanning and comparison with Uni-Navid.}
    \label{fig:bass}
    \vspace{-10pt}
\end{figure}

We evaluate BASS on random trajectories across GRScenes. Figures~\ref{fig:navigation} and \ref{fig:bass}(a-d) demonstrate our closed-loop execution: the robot smoothly tracks subgoals in simple layouts, and dynamically replans recovery routes when BPL detects physical deviations exceeding $D_{\text{thres}}$ in complex topologies. Furthermore, we compare ULVN against Uni-Navid, a state-of-the-art end-to-end model (Fig.~\ref{fig:bass}e). While Uni-Navid's purely reactive, egocentric navigation suffers severe trajectory oscillations, ULVN executes a smooth, efficient trajectory guided by its topological graph. This confirms that lightweight structural memory is crucial for mitigating the myopic planning inherent in purely reactive methods.

Table~\ref{tab:navigation} quantitatively compares our framework against end-to-end baselines and ablates local planner integrations. ULVN significantly outpaces purely reactive models like Uni-Navid and UniGoal, achieving a peak Success Rate (SR) of $71.9\%$ with NoMaD and maintaining a substantially higher SPL. We also ablate the temporal distance loss during the local planner's training (denoted as \textit{-A}). Removing this loss severely degrades performance, particularly for the regression-based ViNT, whereas the generative NoMaD is comparatively more robust. This reveals that temporal distance acts as a crucial auxiliary training signal. However, replacing our BPL filter with temporal-distance-based localization during execution (\textit{ULVN+ViNT w. $d_{temp}$}) reduces SR to $59.6\%$, proving that while temporal distance benefits local control training, our VPR-based BPL is strictly necessary for reliable global localization.

We observe a clear performance gap between localization accuracy ($\sim$95\%) and navigation Success Rate ($\sim$71\%). This gap is primarily due to the physical limitations of the local planners. Specifically, limited obstacle perception and oscillatory trajectory generation, which can occasionally cause the robot to fail despite possessing the correct topological subgoal.

\subsection{Real-world Evaluation}

To validate sim-to-real transfer, we deployed ULVN on a Diablo wheeled robot equipped with an Azure Kinect camera and NVIDIA Jetson Orin. Figure~\ref{fig:realworld} illustrates a representative trial. The initial planned route required a right turn, but physical disturbances caused the robot to drift off-path. Because BPL continuously tracks the observation against the graph, the entropy-adaptive filter concentrates belief on an off-route node. Once the topological distance between the MAP node and the target subgoal exceeded our deviation threshold, BASS successfully triggered a real-time replanning event, issuing new subgoals that safely guided the robot to the destination. This confirms the framework's practical viability for robust, odometry-free physical deployment.

\section{Conclusion}
\label{sec:conclusion}

We present ULVN, an image-goal navigation framework operating entirely from unordered RGB image collections, eliminating the need for odometry or temporal priors. To resolve severe perceptual aliasing, our RAVEL pipeline constructs a robust topological skeleton by reconciling local geometric verification with global structural coherence. On this graph, our Belief Propagation Localization (BPL) filter employs entropy-adaptive fusion to maintain reliable tracking despite severe visual noise, structural occlusions, and complex loops. Supported by this state estimation, our Belief-Aware Subgoal Search (BASS) enables consistent closed-loop execution with dynamic drift recovery. Extensive sim-to-real evaluations demonstrate that our mapping-then-navigating paradigm is highly resilient to perceptual degradation and overcomes the myopic trajectory oscillations inherent in purely reactive end-to-end models. Ultimately, ULVN validates lightweight topological memory as a highly scalable, robust solution for generalizable navigation in unstructured environments.

\section*{Acknowledgement.}
This work was supported by the National Natural Science Foundation of China (U22A2095).

\bibliographystyle{splncs04}
\bibliography{main}

\clearpage

\appendix

\section{Appendix Overview}

This supplementary material provides comprehensive implementation details, mathematical derivations, and extended experimental analysis to support the main paper. The content is organized as follows:

\begin{itemize}
    \item \textbf{Section~\ref{sec:data_collection}: Data Collection Details.} We elaborate on the pipeline used to generate the ground-truth topological graphs and the unordered image dataset from simulation environments.

    \item \textbf{Section~\ref{sec:ravel_details}: RAVEL Extended Algorithmic Details.} We provide an in-depth explanation of the Robust Augmentation and VErification of Landmarks (RAVEL) framework, including the data-driven threshold calibration and the structural pruning mechanisms.

    \item \textbf{Section~\ref{sec:bpl_derivation}: Belief Propagation Localization (BPL) Derivation.} This section offers the mathematical formulation behind our localization module, detailing the graph-based transition model and the entropy-based adaptive fusion strategy.

    \item \textbf{Section~\ref{sec:bass_details}: Belief-Aware Subgoal Search (BASS) Details.} We describe the complete workflow of our navigation policy, specifically focusing on the dynamic subgoal selection and the deviation-triggered replanning logic.

    \item \textbf{Section~\ref{sec:extended_exp}: Extended Experimental Results.} We present additional quantitative ablation studies on graph construction and localization robustness, alongside qualitative visualizations of navigation trajectories in complex scenarios.

\end{itemize}

\section{Data Collection Details}
\label{sec:data_collection}
This section provides an expanded and more detailed description of the acquired dataset and the data acquisition
pipeline used to construct both the unordered landmark library and the corresponding
ground-truth topological structures for ULVN. While the main paper introduces each step
in a concise manner, here we aim to offer a clearer narrative explaining why each
transformation is necessary, how the intermediate representations are constructed, and
how they ultimately contribute to the reliable evaluation of mapping, localization, and
navigation. The entire pipeline converts a raw simulated scene into a sparse skeleton
graph and a set of camera observations, which together form the reference topology.

\subsection{Acquired Dataset Information}
The dataset construction utilized a spatial discretization strategy based on the skeletal structure of the environment. Sampling nodes were systematically generated from the skeleton map and stored as directional pose vectors $[x, y, \text{yaw}]$, which subsequently guided the image acquisition process. The data collection spanned ten distinct environments, divided equally into five Home and five Commercial settings to ensure environmental diversity. Specifically, the image distribution for the residential scenarios Home 0, 1, 2, 3, and 4 comprises 540, 226, 332, 216, and 342 frames, respectively. Correspondingly, the commercial scenarios Commercial 0, 1, 2, 3, and 4 contributed 474, 398, 298, 388, and 380 frames, respectively. In total, the dataset consists of 3,594 high-resolution images with dimensions of $1920 \times 1080$ pixels, providing a dense and high-quality visual representation for topological graph construction.

\subsection{Data Collection Pipeline Details}
Our pipeline roughly follows a sequence of stages: extracting a 3D occupancy grid,
compressing it to a 2D traversability map, computing an ESDF, detecting medial-axis
regions, generating a thinned skeleton, sampling topological nodes, building a
connectivity graph, and finally capturing RGB images at node locations. Below, we
describe these stages.

\vspace{8pt}
\noindent\textbf{3D Occupancy Extraction.}
We begin by exporting a voxelized 3D occupancy grid from the simulator. Each voxel in this grid is recorded as
\begin{equation}
O(x,y,z)\in\{0,1\},
\end{equation}
where an occupied voxel corresponds to any static object or structural element in the
scene. Since the robot operates on a ground plane and its sensors only perceive a
limited vertical band, we restrict our attention to a height slice
\begin{equation}
z\in[z_{\min}, z_{\max}],
\end{equation}
using values determined by the robot design:
\begin{equation}
z_{\min}=\text{0.2\,m}, \qquad
z_{\max}=\text{1.2\,m}.
\end{equation}
The voxel resolution (e.g., \texttt{0.05\,m}) defines the fidelity of this
representation. This 3D occupancy map serves as the geometric foundation from which all
downstream topological structures are derived.

\vspace{8pt}
\noindent\textbf{2D Traversability Projection.}
To produce a representation compatible with planar navigation, we collapse the selected
height slice into a 2D grid:
\begin{equation}
G(x,y)=\max_{z\in[z_{\min},z_{\max}]} O(x,y,z).
\end{equation}
A cell is considered blocked if \emph{any} voxel in the vertical interval is occupied,
ensuring conservative treatment of overhanging obstacles. The resulting 2D discrete map
uses a resolution of \texttt{0.05\,m}. This representation encodes
whether each position on the floor plane is traversable, and it forms the basis for
distance computation and skeleton extraction.

\vspace{8pt}
\noindent\textbf{ESDF Computation.}
To reason about free-space geometry in a continuous and navigation-friendly manner, we
compute an Euclidean Signed Distance Field (ESDF) on the obstacle map. For each cell:
\begin{equation}
\mathrm{ESDF}(x,y)=
\begin{cases}
+d(x,y), & G(x,y)=0,\\[2pt]
-d(x,y), & G(x,y)=1,
\end{cases}
\end{equation}
where $d(x,y)$ is the Euclidean distance to the nearest obstacle cell. This computation
is performed using the Euclidean distance transform algorithm (implemented via \texttt{scipy.ndimage}). The ESDF not only supports robust free-space queries
but also reveals geometric structures that guide the extraction of medial-axis curves.

\vspace{8pt}
\noindent\textbf{Gradients and Flux for Medial-Axis Detection.}
To identify points lying near the centers of corridors or open regions, we compute the
ESDF gradient:
\begin{equation}
\nabla\mathrm{ESDF}(x,y)=
\left(
\frac{\partial \mathrm{ESDF}}{\partial x},
\frac{\partial \mathrm{ESDF}}{\partial y}
\right),
\end{equation}
using finite difference for stability. We then evaluate the
gradient flux,
\begin{equation}
F(x,y)=-\nabla\cdot\left(
\frac{\nabla\mathrm{ESDF}(x,y)}
     {\|\nabla\mathrm{ESDF}(x,y)\|}
\right),
\end{equation}
which highlights locations where distance gradients converge---a characteristic of
topological ``centerlines.'' Applying a threshold
\begin{equation}
F(x,y)>\tau_{\mathrm{flux}},\qquad
\tau_{\mathrm{flux}}=\text{$-0.01$},
\end{equation}
yields a coarse medial-axis mask that roughly traces the connectivity of the navigable
space.

\begin{algorithm}[t]
\caption{Data Collection Pipeline: Skeleton Extraction}
\label{alg:skeleton_extraction}
\begin{algorithmic}[1]
\Require 3D Simulator Environment, Height thresholds $[z_{\min}, z_{\max}]$
\Ensure Skeleton Map $\mathcal{S}_{\text{GT}}$

\State \textbf{Phase 1: Environmental Voxelization}
\State Initialize 3D occupancy grid $\mathcal{O}$
\For{each voxel $v=(x, y, z)$ in environment}
    \State $\mathcal{O}[x, y, z] \gets 1$ \textbf{if} obstacle \textbf{else} $0$
\EndFor

\State \textbf{Phase 2: 2D Projection}
\State Initialize 2D grid $\mathcal{G}_{2D}$
\For{each cell $(x, y)$}
    \State $\mathcal{G}_{2D}[x, y] \gets \max_{z \in [z_{\min}, z_{\max}]} \mathcal{O}[x, y, z]$ \Comment{Flatten 3D map}
\EndFor

\State \textbf{Phase 3: ESDF Computation}
\State Compute Euclidean distance field $d(x,y)$ for $\mathcal{G}_{2D}$
\For{each cell $(x, y)$}
    \If{$\mathcal{G}_{2D}[x, y] = 1$}
        \State $\text{ESDF}[x, y] \gets -d(x, y)$ \Comment{Negative inside obstacles}
    \Else
        \State $\text{ESDF}[x, y] \gets d(x, y)$ \Comment{Positive in free space}
    \EndIf
\EndFor

\State \textbf{Phase 4: Medial-Axis Detection}
\For{each cell $(x, y)$ in ESDF}
    \State Compute gradient $\mathbf{g} \gets \nabla \text{ESDF}[x, y]$
    \State Compute flux $F(x, y)$ based on $\mathbf{g}$ divergence
    \If{$F(x, y) > \tau_{\text{flux}}$}
        \State $\mathcal{S}_{\text{raw}}[x,y] \gets 1$ \Comment{Mark as candidate skeleton}
    \EndIf
\EndFor

\State \textbf{Phase 5: Pruning and Thinning}
\State $\mathcal{S} \gets \text{MorphologicalThinning}(\mathcal{S}_{\text{raw}})$ \Comment{Thin to 1-pixel width}
\For{each branch $b \in \mathcal{S}$}
    \If{$\text{Length}(b) < L_{\min}$}
        \State Remove branch $b$ from $\mathcal{S}$ \Comment{Prune short artifacts}
    \EndIf
\EndFor

\State \Return $\mathcal{S}$
\end{algorithmic}
\end{algorithm}

\begin{algorithm}[t]
\caption{Data Collection Pipeline: Image Capture \& Node Sampling}
\label{alg:node_sampling}
\begin{algorithmic}[1]
\Require Skeleton Map $\mathcal{S}$, Node Spacing $s$, Max Edge Distance $d_{\text{max}}$
\Ensure Topological Graph $\mathcal{G}_{\text{GT}}$, Image Dataset $\mathcal{I}$

\State \textbf{Phase 1: Node Sampling on Skeleton}
\State $V_{\text{sampled}} \gets \emptyset$
\For{each path segment in $\mathcal{S}$}
    \State Sample nodes at regular interval $s$
    \State $V_{\text{sampled}} \gets V_{\text{sampled}} \cup \{ \text{sampled nodes} \}$
\EndFor
\State $V \gets \text{NonMaxSuppression}(V_{\text{sampled}}, r_{\text{nms}})$ \Comment{Remove duplicate nodes}

\State \textbf{Phase 2: Graph Connectivity}
\State $E \gets \emptyset$
\For{each pair $v_i, v_j \in V$}
    \If{$\text{dist}(v_i, v_j) < d_{\text{max}}$ \textbf{and} $\text{LineOfSight}(v_i, v_j)$}
        \State $E \gets E \cup \{(v_i, v_j)\}$ \Comment{Connect navigable nodes}
    \EndIf
\EndFor
\State $\mathcal{G}_{\text{GT}} \gets (V, E)$

\State \textbf{Phase 3: Image Acquisition \& Orientation}
\State Initialize dataset $\mathcal{I} \gets \emptyset$
\For{each node $v_i \in V$}
    \State \textit{// Orientation via PCA or Gradient}
    \State $\mathbf{g}_i \gets \nabla \text{ESDF}(v_i)$
    \State $\theta_i \gets \text{ComputePrincipalDirection}(\mathbf{g}_i)$

    \State Move agent to position $v_i$ with orientation $\theta_i$
    \State Capture image $I_i$
    \State Add $(I_i, v_i, \theta_i)$ to $\mathcal{I}$
\EndFor

\State \Return Image set $\mathcal{I}$ with poses
\end{algorithmic}
\end{algorithm}

\vspace{8pt}
\noindent\textbf{Skeleton Extraction and Thinning.}
The raw medial-axis mask tends to be thick and noisy. To obtain a clean, 1-pixel-wide
topological skeleton, we apply several refinement steps: connected-component filtering,
morphological thinning (e.g., Zhang--Suen~\cite{zhang1984fast} or Guo--Hall~\cite{guo1989parallel}), and pruning of insignificant
branches. In particular, any branch shorter than
\begin{equation}
L_{\min}=15\text{ px}
\end{equation}
is removed to avoid spurious nodes and tiny sub-branches that do not meaningfully
reflect environment structure. The thinned skeleton is a compact curve network
representing the core topology of the navigable region.

\vspace{8pt}
\noindent\textbf{Sampling Topological Nodes.}
To convert the skeleton into a graph representation, we sample nodes along its curves.
Uniform sampling with step size
\begin{equation}
s=\text{0.5\,m}
\end{equation}
provides consistent spatial coverage. Because thinning may still produce localized
clusters, we apply Non-Maximum Suppression (NMS) with radius
\begin{equation}
r_{\mathrm{NMS}}=\text{0.5\,m},
\end{equation}
which ensures that nodes are well distributed and non-redundant. We increase sampling
density around junctions (skeleton points with degree $\ge3$), since these locations are
crucial for representing branching structure in the environment.

\vspace{8pt}
\noindent\textbf{Graph Connectivity Recovery.}
Once node positions $\{v_i\}$ are established, edges are inferred by examining local
connectivity along the skeleton. Two nodes are connected if they lie on the same
skeleton branch and their geodesic separation is smaller than a threshold:
\begin{equation}
d(v_i,v_j)<d_{\max},\qquad d_{\max}=\text{1\,m}.
\end{equation}
This process yields the ground-truth topological graph
\begin{equation}
G_{\mathrm{GT}}=(V,E),
\end{equation}
which encodes the essential navigational structure. Edges may optionally store geometric
lengths or other metadata used during evaluation.

\vspace{8pt}

\vspace{8pt}
\noindent\textbf{RGB Image Capture and Camera Orientation.}
To pair topological locations with visual observations, we place the robot's virtual
camera at each node $v_i$ and capture RGB images. In addition to the node positions, our
pipeline also computes a set of canonical viewing orientations at each node by exploiting
the local geometry of the ESDF and its medial-axis skeleton.

For each sampled node position $p_i=(x_i,y_i)$ on the skeleton, we first map it back to
the underlying ESDF grid and query the local ESDF gradient
\begin{equation}
\nabla\mathrm{ESDF}(x_i,y_i)=
\bigl(g_x(x_i,y_i),\, g_y(x_i,y_i)\bigr).
\end{equation}
The gradient direction
\begin{equation}
\theta_{\mathrm{grad}} = \operatorname{atan2}\bigl(g_y(x_i,y_i),\,g_x(x_i,y_i)\bigr)
\end{equation}
points from the skeleton centerline toward nearby obstacles or free-space boundaries.
To align the camera with the corridor direction, we define a local tangent orientation
as a $90^\circ$ rotation of this gradient:
\begin{equation}
\theta_{\mathrm{tan}} = \theta_{\mathrm{grad}} + \frac{\pi}{2},
\end{equation}
which points along the medial axis and thus approximates the direction of travel.

Around this tangent direction we generate a discrete set of headings by adding a
fixed angular offset:
\begin{equation}
\theta_{i,k} = \theta_{\mathrm{tan}} + k\,\Delta\theta,
\qquad k=0,\dots,K-1,
\end{equation}
where $\Delta\theta$ is the angular interval (e.g., $\Delta\theta=60^\circ$ so that
$K=6$ views are uniformly spaced over $360^\circ$). In practice, the positions and
their associated headings are stored as
\begin{equation}
[x_i,\; y_i,\; \theta_{i,0},\; \theta_{i,1},\dots,\theta_{i,K-1}],
\end{equation}
and the simulator places the camera at $(x_i,y_i)$ while yawing it to each
$\theta_{i,k}$ in turn to render multiple views per node. The camera model, its image
resolution $1920\times1080$, and field of view
$121.04^\circ \times 35.20^\circ$ (horizontal $\times$ vertical) follow the robot
configuration used in our experiments. The resulting unordered collection of images
\begin{equation}
\mathcal{L}=\{I_{i,k}\}_{i=1,\dots,N}^{k=0,\dots,K-1}
\end{equation}
constitutes the sole visual input to ULVN and is used to evaluate the mapping and
localization components. In practice, we sample 2 images for every node.

\vspace{8pt}
\noindent\textbf{Summary.}
Overall, this pipeline transforms raw simulation data into a topologically meaningful
graph and a corresponding set of camera observations. Importantly, these ground-truth
structures are used exclusively for evaluation: ULVN itself receives only unordered RGB
images without metric priors, odometry, or temporal information. By standardizing the
topology generation process, our dataset ensures fair comparisons across diverse
environments and provides a robust foundation for assessing the performance of
topological mapping and navigation systems.
All parameters were fixed across all environments and chosen to match the robot footprint / corridor width scale; Section~\ref{sec:extended_exp} reports robustness to key graph-construction hyperparameters.

\section{RAVEL: Extended Algorithmic Details}
\label{sec:ravel_details}
In this section, we provide an expanded explanation of RAVEL, our
\textbf{R}obust \textbf{A}ugmentation and \textbf{VE}rification of \textbf{L}andmarks
pipeline. The main paper introduces RAVEL at a conceptual level; here we offer
greater detail, including motivation, design principles, and algorithmic
steps. Our aim is to clarify how RAVEL transforms a noisy set of unordered
RGB images into a structurally reliable visual topological graph.

Constructing a graph purely from RGB images without temporal or metric
priors is inherently challenging. Direct appearance-based retrieval often
introduces false positives when visually similar but spatially distant
locations match. Conversely, strict geometric thresholds may eliminate
valid edges, fragmenting the topology. RAVEL addresses these issues by
calibrating the verification process to the statistics of the environment
and then enforcing global structural consistency through principled pruning
and loop reinsertion. The following sections describe these components.

\vspace{6pt}
\noindent\textbf{Global Retrieval and Candidate Generation.}
Given an unordered image library $L=\{I_1,\ldots,I_N\}$, each image is first
mapped to a global descriptor:
\begin{equation}
z_i = f(I_i), \qquad z_i \in \mathbb{R}^D.
\end{equation}
Descriptors are L2-normalized and indexed using FAISS. Their purpose is not
to make final decisions about graph connectivity, but to provide an efficient
way to shortlist potentially related images. For each image, we retrieve all
neighbors within a descriptor distance threshold $d_{\mathrm{VPR}}$, which will be
determined automatically in the calibration stage. This retrieval expands
the search space only where warranted by descriptor similarity, thereby
reducing the number of expensive geometric verifications.

\vspace{6pt}
\noindent\textbf{Motivation for Data-Driven Calibration.}
The performance of geometric verification depends critically on two
quantities: the inlier threshold $\tau$ defining whether two images share a
geometrically valid relation, and the descriptor cutoff $d_{\mathrm{VPR}}$ for
candidate retrieval. In classical pipelines, these thresholds are fixed
constants that must be manually tuned for each environment. However,
illumination, viewpoint, texture distribution, and scene layout vary widely
across environments, making universal fixed thresholds unstable.

RAVEL therefore performs a \emph{one-shot calibration} by probing the
matching statistics of the scene itself. This yields thresholds that
automatically adapt to the difficulty and texture richness of the
environment.

\vspace{6pt}
\noindent\textbf{Anchor-Based Probing and Score Collection.}
To estimate meaningful thresholds, we select two anchor images: a base
anchor $q_0$ and its farthest descriptor neighbor
\begin{equation}
q_f = \arg\max_j\|z_{q_0}-z_j\|_2.
\end{equation}
These anchors collectively span both visually similar and visually diverse
regions of the environment. For each anchor, we exhaustively match it with
all images in $L$, obtaining pairs of measurements:
\begin{equation}
(c_{qj}, d_{qj}) =
(\text{inlier count via LightGlue+RANSAC}, \ \|z_q - z_j\|_2).
\end{equation}
These samples reflect the scene's own distribution of match quality.

\vspace{6pt}
\noindent\textbf{Cluster-Based Threshold Derivation.}
RAVEL pools the inlier counts of both anchors and performs two-means
clustering. The high-confidence cluster $\mathcal{S}_h$ represents strong
geometric matches, while $\mathcal{S}_l$ includes weaker or spurious ones.
We define the inlier threshold as a midpoint between the lowest strong
match and the highest weak match:
\begin{equation}
\tau =
\frac{1}{2}\left(\min(\mathcal{S}_h)+\max(\mathcal{S}_l)\right).
\end{equation}
Likewise, the descriptor threshold is chosen as the largest descriptor
distance among strong matches:
\begin{equation}
d_{\mathrm{VPR}} =
\max_{(q,j)\in\mathcal{S}_h} d_{qj}.
\end{equation}
This calibration technique reliably separates meaningful relations from
noise without any environment-specific hyperparameter tuning.

\vspace{6pt}
\noindent\textbf{Geometric Verification and Graph Assembly.}
For every candidate pair $(i,j)$ retrieved by the calibrated descriptor
threshold, we perform geometric verification. LightGlue extracts local
features and produces correspondences, followed by a RANSAC-based test
(e.g., homography or essential matrix consistency). The resulting inlier
count $c_{ij}$ determines whether an edge is formed:
\begin{equation}
\label{eq:25}
W_{ij}=W_{ji}=
\begin{cases}
c_{ij}, & c_{ij}>\tau,\\[2pt]
0, & \text{otherwise}.
\end{cases}
\end{equation}
This yields an initially dense graph that may contain cycles, shortcuts,
and occasional false positives--a typical outcome when geometric signals
are noisy or repetitive textures exist.

\vspace{6pt}
\noindent\textbf{Need for Structural Pruning.}
Local verification alone cannot guarantee a coherent global structure. For
example, highly textured regions may produce many strong-but-spurious
connections, forming cycles that do not correspond to true spatial
traversability. Conversely, strict local cuts may omit edges that are
important for maintaining connectivity. To resolve these conflicts, RAVEL
uses a Maximum Spanning Forest (MSF) to form a global structural backbone.

\vspace{6pt}
\noindent\textbf{Maximum Spanning Forest (MSF).}
We compute an MSF using Kruskal's algorithm. The MSF:
\begin{itemize}
    \item retains the strongest verified edges,
    \item eliminates all cycles,
    \item ensures connectivity within each component,
    \item promotes global structural coherence.
\end{itemize}
Intuitively, this process finds the minimum number of edges needed to
connect the graph while preserving only the most reliable relations.

\vspace{6pt}
\noindent\textbf{Loop Reinsertion.}
Real environments often contain genuine loops---for instance, rectangular
corridors or circular paths. Since MSF removes all cycles, RAVEL reintroduces
only those edges whose geometric support is exceptionally strong. An edge
$(i,j)$ is restored if
\begin{equation}
W_{ij} > \tau_{\mathrm{add}},
\qquad
\tau_{\mathrm{add}} = 1.5\,\tau.
\end{equation}
By reinserting only highly confident loops, RAVEL strikes a balance between
topological sparsity and representational completeness.

\vspace{6pt}
\begin{algorithm}[t]
\caption{RAVEL Graph Construction}
\label{alg:ravel_construction}
\begin{algorithmic}[1]
\Require Unordered image set $\mathcal{I} = \{I_1,\dots,I_N\}$
\Ensure Topological Graph $\mathcal{G}=(\mathcal{V}, \mathcal{E}, W)$

\State \textbf{Phase 1: Initialization \& Feature Extraction}
\State Compute global descriptors $\mathcal{Z} = \{z_i \mid z_i=f(I_i)\}$
\State Build FAISS index using $\mathcal{Z}$ for efficient retrieval

\State \textbf{Phase 2: Parameter Estimation}
\State Select anchors $q_0$ (random) and $q_f$ (farthest neighbor of $q_0$)
\For{$q \in \{q_0, q_f\}$}
    \State Compute geometric inliers $c_{qj}$ for all $j$ \Comment{Using LightGlue + RANSAC}
    \State Record descriptor distances $d_{qj}$
\EndFor
\State Cluster inlier statistics into sets $\mathcal{S}_{high}$ and $\mathcal{S}_{low}$
\State Derive verification threshold $\tau$ and retrieval radius $d_{\mathrm{VPR}}$

\State \textbf{Phase 3: Dense Graph Construction}
\State Initialize adjacency matrix $W \gets \mathbf{0}_{N \times N}$
\For{each image $i \in \{1, \dots, N\}$}
    \State Retrieve candidates $\mathcal{C}_i \gets \{j \mid \|z_i-z_j\| \le d_{\mathrm{VPR}}\}$ \Comment{FAISS Radius Search}
    \For{each $j \in \mathcal{C}_i$}
        \State $c_{ij} \gets \text{GeometricVerify}(I_i, I_j)$
        \If{$c_{ij} > \tau$}
            \State $W_{ij} \gets c_{ij}$ \Comment{Edge weight is inlier count}
        \EndIf
    \EndFor
\EndFor

\State \textbf{Phase 4: Topology Refinement}
\State $\mathcal{T}_{\text{MSF}} \gets \text{MaximumSpanningForest}(W)$ \Comment{Ensures global connectivity}
\State $E_{final} \gets \mathcal{T}_{\text{MSF}} \cup \{ (i,j) \mid W_{ij} > \tau_{\mathrm{add}} \}$ \Comment{Re-insert strong loop closures}
\State Construct final graph $\mathcal{G} = (V, E_{final}, W)$

\State \Return $\mathcal{G}$
\end{algorithmic}
\end{algorithm}

\vspace{6pt}
\noindent\textbf{Discussion and Remaining Challenges.}
RAVEL significantly improves edge precision and global coherence relative to
retrieval-only or naive verification pipelines. Nevertheless, extremely
texture-poor environments or large open areas may still introduce ambiguity.
These cases reflect fundamental limitations of monocular imagery rather than
the algorithm itself. Despite such challenges, RAVEL consistently provides a
stable foundation for the localization and navigation modules described in the
main paper.

\section{Belief Propagation Localization: Additional Derivation}
\label{sec:bpl_derivation}
\begin{algorithm}[t]
\caption{Belief Propagation Localization Update}
\label{alg:bpl_update}
\begin{algorithmic}[1]
\Require Current Observation $I_t$, Previous Belief $\mathbf{b}_{t-1}$, Transition Matrix $\mathbf{T}$, Node Descriptors $\mathcal{Z}=\{z_1, \dots, z_N\}$
\Ensure Updated Belief State $\mathbf{b}_t$

\State \textbf{1. Belief Prediction (Transition)}
\State $\bar{\mathbf{b}}_t \gets \mathbf{b}_{t-1} \cdot \mathbf{T}$ \Comment{Propagate belief via graph topology}

\State \textbf{2. Observation Likelihood Estimation}
\State Extract feature vector $z_t \gets f(I_t)$
\For{each node $v_i \in V$}
    \State $d_i \gets \|z_t - z_i\|_2^2$ \Comment{Euclidean distance}
    \State $L(v_i | I_t) \gets \exp\left( -\lambda \cdot d_i \right)$ \Comment{Compute Likelihood}
\EndFor

\State \textbf{3. Adaptive Weight Calculation}
\State $\eta_t \gets \dfrac{-\sum_i \bar{\mathbf{b}}_t(v_i)\log \bar{\mathbf{b}}_t(v_i)}{\log N}$ \Comment{Normalized entropy}
\State Compute $w_p(\eta_t)$ using Eq.~(\ref{eq:25})
\State $w_o(\eta_t) \gets 1 - w_p(\eta_t)$

\State \textbf{4. Fusion and Update}
\For{each node $v_i \in V$}
    \State $\mathbf{b}_t(v_i) \gets \bar{\mathbf{b}}_t(v_i)^{w_p} \cdot L(v_i | I_t)^{w_o}$ \Comment{Adaptive Fusion}
\EndFor

\State \textbf{5. Normalization}
\State $\eta \gets \sum_{j} \mathbf{b}_t(v_j)$
\State $\mathbf{b}_t \gets \mathbf{b}_t / \eta$ \Comment{Ensure $\sum \mathbf{b}_t = 1$}

\State \Return $\mathbf{b}_t$
\end{algorithmic}
\end{algorithm}

This section provides a deeper understanding of the belief propagation (BP) localization method used in ULVN. While the main paper outlines the basic principles of BP localization, here we extend the derivation by discussing the underlying mathematical framework, explaining the topological transition model, and providing insights into how belief propagation is performed. We also describe the adaptive fusion process that combines predicted beliefs with new observations.

\textbf{Motivation for Belief Propagation Localization.}
The goal of localization is to estimate the robot's position within the topological graph. This is done by maintaining a belief distribution over all the possible nodes, which represents the probability of the robot being at each location. Traditional localization methods often rely on sequential filtering (such as Kalman filters) that depend heavily on temporal priors. However, our method has no direct temporal or odometric information about robot movement. As a result, we must rely on the topology of the environment and adaptively update our belief based on new observations.

The belief propagation method that we propose solves this problem by propagating the belief across the graph structure and incorporating new observations through an iterative process. It does not assume sequential data or odometry, and instead relies on the spatial relationships encoded in the topological graph to update the belief at each step.

\vspace{8pt}
\noindent\textbf{Graph Representation of Localization.}
Our localization approach works on a topological graph \(G = (V, E)\), where \(V\) represents the set of nodes (locations in the environment), and \(E\) represents the edges (connections between locations). Each node \(v_i \in V\) corresponds to a unique landmark (or a set of landmarks), and the edges \(E\) represent the navigability between these landmarks.

Let the belief state \(b_t \in \mathbb{R}^N\) represent the probability distribution over the \(N\) nodes at time step \(t\). This belief is initially uniform or initialized based on the first retrieval, and is updated by combining prior knowledge from the graph and new observations from the robot's sensors.

\vspace{8pt}
\noindent\textbf{Belief Prediction via Multi-Step Transition.}
The belief prediction step utilizes a transition matrix \(\mathbf{T}\) derived from the graph's adjacency matrix \(\mathbf{A}\). Unlike simple Markov models that only consider immediate neighbors, our implementation accounts for reachability up to \(K\) steps (where \(K=3\) in our experiments) to model the potential motion of the robot between updates.

We construct a cumulative reachability matrix by summing powers of the adjacency matrix:
\begin{equation}
\mathbf{C} = \sum_{k=0}^{K} \mathbf{A}^k,
\end{equation}
where \(\mathbf{A}^0 = \mathbf{I}\) (identity matrix). The transition matrix \(\mathbf{T}\) is obtained by row-normalizing \(\mathbf{C}\) to ensure stochasticity:
\begin{equation}
\mathbf{T}_{ij} = \frac{\mathbf{C}_{ij}}{\sum_{j} \mathbf{C}_{ij}}.
\end{equation}
To predict the robot's belief state at time \(t\), we propagate the previous posterior belief \(b_{t-1}\) through this transition model:
\begin{equation}
\bar{b}_t = b_{t-1} \cdot \mathbf{T}.
\end{equation}
Here, \(\bar{b}_t\) represents the prior belief at time \(t\) before accounting for the current observation. This formulation effectively diffuses the probability mass to the topological neighborhood of the previous estimate.

\vspace{8pt}
\noindent\textbf{Adaptive Fusion of Belief and Observation.}
Once the belief has been propagated, we fuse it with the current visual observation \(I_t\).

\textit{1. Observation Likelihood:}
We compute the likelihood \(L(v_i|I_t)\) based on the Euclidean distance \(d(z_t, z_i)\) between the query embedding \(z_t\) and the node embeddings \(z_i\). To handle varying dynamic ranges in feature space, we calculate an adaptive scaling factor \(\lambda\):
\begin{equation}
\lambda = \frac{\ln(\delta)}{q_{0.975}(\mathbf{d}) - q_{0.025}(\mathbf{d})},
\end{equation}
where \(\mathbf{d}\) is the vector of distances between \(z_t\) and all nodes, and \(q\) denotes the quantile function. The likelihood is then modeled as:
\begin{equation}
L(v_i|I_t) \propto \exp\left( -\lambda \cdot d(z_t, z_i) \right).
\end{equation}

\textit{2. Entropy-based Weighting:}
A core contribution of our method is the adaptive weighting of prediction versus observation based on the uncertainty of the current belief. We quantify uncertainty using the normalized Shannon entropy \(\eta\) of the belief distribution:
\begin{equation}
\eta(b) = \frac{-\sum_{i} b(v_i) \ln(b(v_i))}{\ln(N)}.
\end{equation}
We define a prediction weight \(w_p\) and an observation weight \(w_o\) using a piecewise linear function based on \(\eta\). When entropy is low (high confidence), we rely more on the motion prediction; when entropy is high (high uncertainty), we rely more on the current observation to relocalize.

Specifically, the prediction weight \(w_p\) is calculated as:
\begin{equation}
w_p(\eta) =
\begin{cases}
0.6 & \text{if } \eta < 0.3 \quad \\
0.6 - 0.75(\eta - 0.3) & \text{if } 0.3 \le \eta \le 0.7 \quad \\
0.3 & \text{if } \eta > 0.7 \quad
\end{cases}
\end{equation}
The observation weight is set complementarily as \(w_o(\eta) = 1.0 - w_p(\eta)\).

\textit{3. Geometric Fusion Update:}
Finally, the updated posterior belief \(b_t\) is computed via the geometric weighted average of the prior belief and the likelihood:
\begin{equation}
b_t(v_i) \propto \bar{b}_t(v_i)^{w_p} \cdot L(v_i | I_t)^{w_o}.
\end{equation}
This multiplicative fusion (equivalent to a weighted sum in log-space) is sharper than additive fusion and effectively suppresses nodes that are not supported by both the transition model and the visual evidence.

\vspace{8pt}
\noindent\textbf{Summary.}
Belief propagation localization provides an elegant solution for estimating the robot's position in a topological map without relying on temporal or odometric data. By combining multi-step topological propagation with entropy-aware adaptive fusion, the system can effectively handle uncertainties. The dynamic weights \(w_p\) and \(w_o\) ensure that the system holds its course when confident but remains responsive to new visual evidence when the location is ambiguous.

\section{Belief-Aware Subgoal Search Details}
\label{sec:bass_details}
\begin{algorithm}[t]
\caption{Belief-Aware Subgoal Search (BASS)}
\label{alg:bass}
\begin{algorithmic}[1]
\Require Topological Map $\mathcal{G}=(\mathcal{V}, \mathcal{E})$, Target Image $I_{goal}$, Look-ahead window $\Delta$
\Ensure Robot trajectory to goal

\State \textbf{Phase 1: Initialization}
\State Initialize belief $b_0$ using BPL given initial observation
\State $v_{goal} \gets \operatorname*{argmax}_{v \in \mathcal{V}} P(v \mid I_{goal})$ \Comment{Goal Identification}
\State $v_{curr} \gets \operatorname*{argmax}_{v \in \mathcal{V}} b_0(v)$
\State $P \gets \text{GlobalPlan}(\mathcal{G}, v_{curr}, v_{goal})$ \Comment{Compute global path $P=\{v_0, \dots, v_{goal}\}$}

\State \textbf{Phase 2: Navigation Loop}
\While{robot has not reached $v_{goal}$}
    \State Obtain current observation $o_t$
    \State $b_t \gets \text{BPL}(b_{t-1}, o_t)$ \Comment{Update Belief State}

    \State \textit{// Deviation Check and Replanning}
    \State $p_{on\_path} \gets \max_{v \in P} b_t(v)$
    \If{$p_{on\_path} < \tau_{threshold}$ \textbf{or} IsDeviated($b_t, P$)}
        \State $v_{curr} \gets \operatorname*{argmax}_{v \in \mathcal{V}} b_t(v)$
        \State $P \gets \text{GlobalPlan}(\mathcal{G}, v_{curr}, v_{goal})$ \Comment{Replan path}
    \EndIf

    \State \textit{// Path-Dependent Subgoal Selection}
    \State $k^* \gets \operatorname*{argmax}_{k} \{ b_t(v_k) \mid v_k \in P \}$ \Comment{Locate index on path}
    \State $v_{sub} \gets P_{\min(k^* + \Delta, |P|)}$ \Comment{Look-ahead selection}

    \State \textit{// Closed-Loop Execution}
    \State $a_t \gets \text{LocalPlanner}(v_{sub}, b_t)$
    \State Execute action $a_t$
\EndWhile
\end{algorithmic}
\end{algorithm}

This section provides a detailed explanation of the Belief-Aware Subgoal Search (BASS) algorithm, which plays a critical role in ULVN for visual navigation. BASS integrates belief propagation and path planning to enable efficient navigation in uncertain environments, where the robot's belief state is continuously updated as it moves. While the main paper introduces BASS at a high level, this section delves into the algorithmic details, providing a step-by-step explanation of its operation and the underlying motivations.

\textbf{Motivation for BASS.}
In environments where a robot must navigate based on visual cues and an uncertain belief state, a simple approach to path planning is not sufficient. Traditional methods typically focus on planning from the robot's current state to a goal, assuming a known map and precise position. However, in visual navigation, where the robot's belief about its position is uncertain and constantly evolving, we need a more robust approach that considers both the belief state and the goal in the context of topological navigation.

BASS addresses this challenge by introducing the concept of \textbf{belief-aware subgoal search}. Instead of rigidly following a sequence of nodes, BASS dynamically identifies immediate subgoals from a global plan that correspond to the robot's current belief. This ensures that the robot targets a reachable, forward-looking waypoint on the path, or detects when it has deviated sufficiently to require replanning.

\vspace{8pt}
\noindent\textbf{The BASS Planning and Execution Process.}
The BASS algorithm operates on the topological map constructed by ULVN. The process is divided into global path planning and dynamic subgoal selection:

1. Initial Localization: The robot's belief is first updated using the Belief Propagation Localization (BPL) module to determine the initial probability distribution over the topological graph.

2. Goal Identification: The goal is represented by a target image, and the robot's belief state is updated to reflect the most likely node that corresponds to the goal. This goal node serves as the destination for the navigation task.

3. Global Path Planning: Before selecting immediate subgoals, a global topological path $P = \{v_0, v_1, \dots, v_{goal}\}$ is computed from the most likely starting node to the goal node. This path consists of a sequence of visual landmarks (nodes) connected by traversable edges.

4. Dynamic Subgoal Selection: As the robot navigates, BASS uses the current belief state to determine the robot's progress along the path $P$. It selects a "look-ahead" node on the path as the immediate subgoal to drive efficient movement.

5. Deviation Check and Replanning: Crucially, BASS continuously verifies if the robot's current belief state is consistent with the nodes contained in the global path $P$. If the robot localizes itself off the path, the system triggers a replanning procedure.

\vspace{8pt}
\noindent\textbf{Path-Dependent Subgoal Selection Mechanism.}
The core of BASS is the mechanism for selecting the immediate target image (subgoal) given the global plan $P$ and the current belief $b_t$. Unlike information-seeking exploration, this process focuses on efficient path traversal.

1. Localization on Path: Let the global path be defined as a sequence of nodes $P$. At time step $t$, the robot estimates its location index $k$ on the path such that the belief $b_t(v_k)$ is maximized among all nodes in $P$.
\begin{equation}
k^* = \operatorname{argmax}_{k} \, b_t(v_k), \quad \text{where } v_k \in P
\end{equation}

2. Look-Ahead Selection: Once the robot's current location index $k^*$ on the path is identified, BASS selects a downstream node as the immediate subgoal to encourage forward progress. The subgoal $v_{sub}$ is chosen as:
\begin{equation}
v_{sub} = v_{k^* + \Delta},
\end{equation}
where $\Delta$ represents a look-ahead window. This ensures the robot moves towards a future waypoint rather than trying to converge on a location it is already passing.

\vspace{8pt}
\noindent\textbf{Closed-Loop Navigation and Replanning.}
Once the subgoals are selected, the robot navigates towards them using a local planner. However, because visual navigation often involves uncertainty in both localization and movement, the robot may deviate from the planned path. In such cases, BASS enables closed-loop navigation by continuously updating the belief state and adjusting the path.

1. Path Execution: The robot follows the path from its current location to the selected subgoal. This path is planned using a local planner that takes into account both the belief state and the topological graph. As the robot moves, its belief about its position is updated using the BPL module.

2. Deviation Detection: If the robot deviates significantly from the planned path, BASS triggers a replanning mechanism. This deviation is detected by comparing the robot's current belief state with the expected belief based on the planned path. If the belief state shows that the robot is no longer on the correct path, a new subgoal is selected, and the robot is reoriented towards it.

3. Replanning and Adaptation: If a deviation occurs or if new information is obtained (e.g., through visual observations), the belief state is updated, and the path to the goal is recomputed. This iterative process continues until the robot reaches the final goal or the belief state converges.

\vspace{8pt}
\noindent\textbf{Computational Complexity and Performance.}
The computational complexity of the BASS algorithm is influenced by the size of the topological graph and the number of subgoals selected. The time complexity for selecting a subgoal involves evaluating all candidate nodes, which requires calculating the expected information gain and belief entropy for each node. This process is \(O(N)\), where \(N\) is the number of nodes in the graph.

Once the subgoal is selected, the path planning step is performed using a graph search algorithm (e.g., Dijkstra or A*), which has a time complexity of \(O(E \log N)\), where \(E\) is the number of edges in the graph. Overall, the complexity of BASS is primarily dominated by the belief propagation and path search steps.

\vspace{8pt}
\noindent\textbf{Summary.}
The Belief-Aware Subgoal Search (BASS) algorithm provides a robust framework for visual navigation by incorporating belief updates and adaptive path planning. By selecting intermediate subgoals based on the robot's belief state, BASS ensures that the robot always moves towards the most informative locations in the environment. This approach is particularly useful in scenarios where the robot's position is uncertain, and it enables efficient navigation in complex, dynamic environments. BASS's closed-loop capabilities allow the robot to recover from deviations and continue to the goal, making it a powerful tool for real-world deployment in visual navigation tasks.

\section{Extended Experimental Results}
\label{sec:extended_exp}
\begin{table*}[ht]
\centering
\caption{Detailed comparison evaluation of different topological graph construction methods on the GRScenes dataset.}
\resizebox{\linewidth}{!}{%
\begin{tabular}{lcccc}
\toprule
Method & Precision & Recall & F1 & Accuracy \\
\midrule
\rowcolor[HTML]{D5D5D5}\multicolumn{5}{c}{\textbf{Commercial}} \\
PlaceNav top 2~\cite{suomela2024placenav} & 0.5078 $\pm$ 0.0453 & 0.6803 $\pm$ 0.0604 & 0.5682 $\pm$ 0.0477 & 0.9943 $\pm$ 0.0011 \\
PlaceNav top 5~\cite{suomela2024placenav} & 0.2324 $\pm$ 0.0262 & 0.7878 $\pm$ 0.0479 & 0.3587 $\pm$ 0.0265 & 0.9857 $\pm$ 0.0022 \\
ViNT top 2~\cite{shah2023vint}     & 0.5291 $\pm$ 0.0724 & 0.6839 $\pm$ 0.1021 & 0.5962 $\pm$ 0.0701 & 0.9951 $\pm$ 0.0013 \\
ViNT top 5~\cite{shah2023vint}     & 0.2823 $\pm$ 0.1686 & \textbf{0.8239} $\pm$ 0.0410 & 0.3882 $\pm$ 0.1522 & 0.9862 $\pm$ 0.0059 \\
RAVEL (Ours) & \textbf{0.7012} $\pm$ 0.0886 & 0.7687 $\pm$ 0.0599 & \textbf{0.7328} $\pm$ 0.0756 & \textbf{0.9972} $\pm$ 0.0011 \\

\midrule
\rowcolor[HTML]{D5D5D5}\multicolumn{5}{c}{\textbf{Home}} \\
PlaceNav top 2~\cite{suomela2024placenav} & 0.5446 $\pm$ 0.0390 & 0.7224 $\pm$ 0.0690 & 0.6203 $\pm$ 0.0478 & 0.9946 $\pm$ 0.0018 \\
PlaceNav top 5~\cite{suomela2024placenav} & 0.3075 $\pm$ 0.1192 & 0.8032 $\pm$ 0.0590 & 0.4149 $\pm$ 0.0491 & 0.9850 $\pm$ 0.0060 \\
ViNT top 2~\cite{shah2023vint}     & 0.4752 $\pm$ 0.1361 & 0.6910 $\pm$ 0.1406 & 0.6062 $\pm$ 0.0857 & 0.9935 $\pm$ 0.0022 \\
ViNT top 5~\cite{shah2023vint}     & 0.1502 $\pm$ 0.0714 & \textbf{0.8281} $\pm$ 0.0672 & 0.3289 $\pm$ 0.1095 & 0.9761 $\pm$ 0.0094 \\
RAVEL (Ours) & \textbf{0.7196} $\pm$ 0.1291 & 0.7625 $\pm$ 0.1217 & \textbf{0.7402} $\pm$ 0.1252 & \textbf{0.9967} $\pm$ 0.0016 \\

\midrule
\rowcolor[HTML]{D5D5D5}\multicolumn{5}{c}{\textbf{All (Home + Commercial)}} \\
PlaceNav top 2~\cite{suomela2024placenav} & 0.5262 $\pm$ 0.0503 & 0.7113 $\pm$ 0.0805 & 0.6043 $\pm$ 0.0606 & 0.9948 $\pm$ 0.0014 \\
PlaceNav top 5~\cite{suomela2024placenav} & 0.2699 $\pm$ 0.0915 & 0.7655 $\pm$ 0.1584 & 0.3731 $\pm$ 0.0244 & 0.9853 $\pm$ 0.0052 \\
ViNT top 2~\cite{shah2023vint}     & 0.5201 $\pm$ 0.0718 & 0.6775 $\pm$ 0.1252 & 0.5815 $\pm$ 0.0724 & 0.9946 $\pm$ 0.0015 \\
ViNT top 5~\cite{shah2023vint}     & 0.2660 $\pm$ 0.1540 & \textbf{0.8246} $\pm$ 0.0560 & 0.3806 $\pm$ 0.1329 & 0.9816 $\pm$ 0.0108 \\
RAVEL (Ours) & \textbf{0.7104} $\pm$ 0.1048 & 0.7656 $\pm$ 0.0905 & \textbf{0.7365} $\pm$ 0.0976 & \textbf{0.9970} $\pm$ 0.0013 \\

\bottomrule
\end{tabular}
}
\label{stab:ravel}
\end{table*}

\begin{table*}[ht]
\centering
\caption{Detailed quantitative evaluation of topological graph construction on the GRScenes dataset.}

\setlength{\tabcolsep}{2pt}
\renewcommand{\arraystretch}{0.9}
\resizebox{\linewidth}{!}{%
\begin{tabular}{llcccc}
\toprule

\textbf{Scenes} & \textbf{Metric} & \textbf{Top-k ANN} & \textbf{Ours w/o (MSF, AVP $\tau$)} & \textbf{Ours w/o MSF} & \textbf{Ours}\\
\midrule
\multirow{4}{*}{\textbf{Home}} & $P$ & 0.1537 $\pm$ 0.0551 & 0.4625 $\pm$ 0.1459 & \textbf{0.7198} $\pm$ 0.1075 & 0.7196 $\pm$ 0.1291 \\
 & $R$ & \textbf{0.8898} $\pm$ 0.1095 & 0.6177 $\pm$ 0.1604 & 0.3521 $\pm$ 0.0492 & 0.7625 $\pm$ 0.1217 \\
 & $F1$ & 0.2598 $\pm$ 0.0835 & 0.5070 $\pm$ 0.0754 & 0.4705 $\pm$ 0.0564 & \textbf{0.7402} $\pm$ 0.1252 \\
 & $Acc.$ & 0.9673 $\pm$ 0.0106 & 0.9929 $\pm$ 0.0020 & 0.9951 $\pm$ 0.0016 & \textbf{0.9967} $\pm$ 0.0016 \\
\midrule
\multirow{4}{*}{\textbf{Commercial}} & $P$ & 0.0953 $\pm$ 0.0341 & 0.2728 $\pm$ 0.1145 & 0.6621 $\pm$ 0.1202 & \textbf{0.7012} $\pm$ 0.0886 \\
 & $R$ & \textbf{0.9343} $\pm$ 0.0274 & 0.8177 $\pm$ 0.0978 & 0.4920 $\pm$ 0.0981 & 0.7687 $\pm$ 0.0599 \\
 & $F1$ & 0.1713 $\pm$ 0.0554 & 0.3922 $\pm$ 0.0967 & 0.5609 $\pm$ 0.0984 & \textbf{0.7328} $\pm$ 0.0756 \\
 & $Acc.$ & 0.9496 $\pm$ 0.0220 & 0.9868 $\pm$ 0.0050 & 0.9962 $\pm$ 0.0011 & \textbf{0.9972} $\pm$ 0.0011 \\
\midrule
\multirow{4}{*}{\textbf{All}} & $P$ & 0.1245 $\pm$ 0.0530 & 0.3676 $\pm$ 0.1590 & 0.6910 $\pm$ 0.1117 & \textbf{0.7104} $\pm$ 0.1048 \\
 & $R$ & \textbf{0.9121} $\pm$ 0.0788 & 0.7177 $\pm$ 0.1637 & 0.4220 $\pm$ 0.1039 & 0.7656 $\pm$ 0.0905 \\
 & $F1$ & 0.2156 $\pm$ 0.0815 & 0.4496 $\pm$ 0.1017 & 0.5157 $\pm$ 0.0894 & \textbf{0.7365} $\pm$ 0.0976 \\
 & $Acc.$ & 0.9584 $\pm$ 0.0188 & 0.9898 $\pm$ 0.0048 & 0.9956 $\pm$ 0.0014 & \textbf{0.9970} $\pm$ 0.0013 \\
\bottomrule
\label{stab:ravel_ablation}
\end{tabular}
}

\end{table*}

This section presents additional experimental results to complement the findings in the main paper.

We provide more detailed results to complement the experiments in the main text. Additionally, we conduct a sensitivity analysis on the adaptive parameters $\tau$ and $d_{\mathrm{VPR}}$ in RAVEL to demonstrate the robustness of the \textit{Anchor selection} strategy and the overall effectiveness of the method. We also analyze the structural properties of the constructed topological graphs by reporting metrics such as the number of connected components, average shortest path length, and graph diameter. Visual comparisons of the topological graphs obtained by different methods are also provided. Considering that most existing methods rely on temporal sequences, we validate our global localization algorithm, BPL, across multiple datasets that including a mount of temporal sequences. To simulate variations in position, orientation, and motion blur that occur during robot movement, we apply techniques such as image cropping, viewpoint modification, and noise injection. These tests further verify the effectiveness and generalization capability of our approach. Finally, we present visualizations of selected navigation trajectories to illustrate the system performance, with a particular focus on demonstrating the efficacy of the replanning mechanism.

\subsection{Experimental Details}
To evaluate the system's performance rigorously, we conducted separate experimental trials for global localization and autonomous navigation using randomly selected trajectories within the mapped environments. For the localization experiments, the testing protocol involved traversing random paths across all scenes to assess position estimation accuracy. Specifically, to ensure statistical significance, five distinct paths were generated for each of the environments. The navigation evaluation followed a similar randomized path generation strategy with slight adjustments to accommodate scene-specific complexities.

\subsection{Extended Experiments}
\noindent\textbf{Evaluation Metrics and Experimental Setup.}
For the extended experiments, we evaluate ULVN across multiple tasks to comprehensively assess its performance. The core tasks include:

1. Topological Mapping:

We validate the robustness of the one-shot calibration mechanism in RAVEL through a sensitivity analysis. Additionally, we statistically evaluate the properties of the constructed topological graphs across multiple dimensions, including the number of connected components, average shortest path length, and graph diameter.

2. Localization:

We conducted simulated global localization experiments on time-series navigation datasets. For each image within a clip, we applied image processing techniques, including cropping, viewpoint modification, and noise injection, to simulate variations in position, orientation, and motion blur encountered during robot motion. The processed images were treated as robot observations, while the original images served as the node images in the topological graph. We performed global localization for each image and reported the resulting accuracy. These experiments were conducted on the RECON \cite{shah2021rapid}, SCAND \cite{karnan2022scand}, GoStanford \cite{hirose2019deep}, and SACSoN \cite{hirose2023sacson} datasets.

3. Navigation:

We present visualizations of navigation paths in selected scenes. Additionally, we provide ablation studies for the navigation method, including experiments on the replanning mechanism and global localization.

\begin{table*}[t]
  \centering
  \caption{Comparison of localization accuracy of different methods on different datasets.}
  \label{stab:localization_accuracy}
  \setlength{\tabcolsep}{8pt}
  \renewcommand{\arraystretch}{0.9}

  \resizebox{\linewidth}{!}{%
    \begin{tabular}{llcccc}
      \toprule
      \multirow{2}{*}{\textbf{Dataset}} & \multirow{2}{*}{\textbf{Condition}} & \multicolumn{4}{c}{\textbf{Localization Accuracy}} \\
      \cmidrule(lr){3-6}
      & & \textbf{MegaLoc}~\cite{berton2025megaloc} & \textbf{JIST}~\cite{berton2023jist} & \textbf{ViNT}~\cite{shah2023vint} & \textbf{BPL (Ours)} \\
      \midrule
      \multirow{4}{*}{\textbf{GoStanford}~\cite{hirose2019deep}}
      & Rotation & \textbf{0.9930} & 0.8047 & 0.9302 & 0.9891 \\
      & Rot + Gauss & 0.8674 & 0.3395 & 0.9295 & \textbf{0.9559} \\
      & Rot + Poisson & 0.8488 & 0.3209 & 0.9226 & \textbf{0.9412} \\
      & Rot + Crop & \textbf{0.9884} & 0.6209 & 0.6140 & 0.9853 \\
      \midrule
      \multirow{4}{*}{\textbf{RECON}~\cite{shah2021rapid}}
      & Rotation & 0.7597 & 0.7394 & 0.9306 & \textbf{0.9429} \\
      & Rot + Gauss & 0.7022 & 0.7039 & 0.9069 & \textbf{0.9375} \\
      & Rot + Poisson & 0.6362 & 0.6937 & 0.9120 & \textbf{0.9062} \\
      & Rot + Crop & 0.6379 & 0.5262 & 0.7733 & \textbf{0.9301} \\
      \midrule
      \multirow{4}{*}{\textbf{SACSoN}~\cite{hirose2023sacson}}
      & Rotation & 0.9507 & 0.6494 & 0.8410 & \textbf{0.9565} \\
      & Rot + Gauss & 0.6669 & 0.3172 & 0.8267 & \textbf{0.8862} \\
      & Rot + Poisson & 0.6089 & 0.3188 & 0.8410 & \textbf{0.8556} \\
      & Rot + Crop & 0.9006 & 0.4356 & 0.5437 & \textbf{0.9394} \\
      \midrule
      \multirow{4}{*}{\textbf{SCAND}~\cite{karnan2022scand}}
      & Rotation & 0.9471 & 0.6960 & 0.8784 & \textbf{0.9753} \\
      & Rot + Gauss & 0.7364 & 0.4486 & 0.8753 & \textbf{0.8772} \\
      & Rot + Poisson & 0.7746 & 0.4481 & 0.8768 & \textbf{0.8827} \\
      & Rot + Crop & 0.8920 & 0.5183 & 0.6305 & \textbf{0.9259} \\
      \midrule
      \textbf{Average} & Overall & 0.8069 & 0.5363 & 0.8270 & \textbf{0.9304} \\
      \bottomrule
    \end{tabular}%
  }
\end{table*}

\vspace{8pt}
\noindent\textbf{Detailed Results of Topological Mapping Performance.}
The detailed data of different topological graph construction methods comparison are presented in Table~\ref{stab:ravel}. A granular examination of the sub-categories reveals that the proposed RAVEL method maintains superior performance stability across diverse environmental contexts. In the Commercial settings, RAVEL achieves a dominant Precision of $0.7012$ and an F1 score of $0.7328$, significantly outpacing the closest competitors. This performance advantage is even more pronounced in the Home scenarios, where RAVEL reaches a Precision of $0.7196$ and an F1 score of $0.7402$. Notably, while baseline methods such as ViNT top 5 experience a drastic reduction in Precision when transitioning from Commercial ($0.2823$) to Home ($0.1502$) environments, RAVEL exhibits remarkable consistency. This suggests that the proposed approach is far more robust to the structural variability and visual clutter variations that distinguish residential layouts from commercial spaces.Regarding statistical variance, the standard deviations reported in the table provide insight into the reliability of the graph construction. While methods like ViNT top 5 achieve the highest Recall values, their high standard deviations in Precision (up to $\pm 0.1686$ in Commercial settings) and extremely low mean Precision scores indicate a reliance on generating excessive false positives, resulting in unstable topological graphs. Conversely, RAVEL demonstrates a balanced trade-off. Although its standard deviation for Precision in Home scenes is higher than in Commercial scenes, reflecting the inherent irregularity of residential environments, the method consistently achieves the highest Accuracy with tight error bounds. This low variance in overall Accuracy confirms that RAVEL offers the most reliable and generalized solution for topological mapping, effectively mitigating the extreme fluctuations observed in heuristic-based baselines.

\begin{figure*}[htbp]
    \centering
    \includegraphics[width=\linewidth]{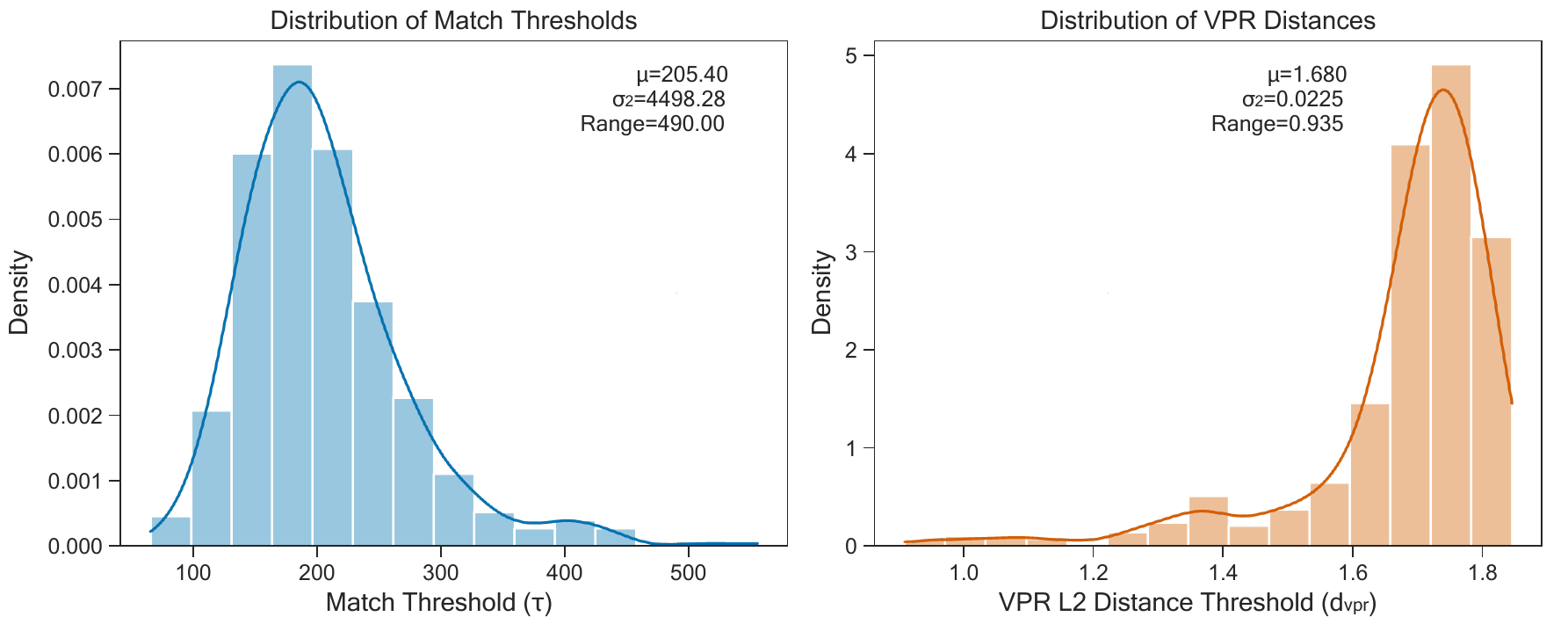}

    \caption{Distribution of adaptive thresholds in RAVEL on the GRScenes dataset.}
    \label{sfig:param_dis}

\end{figure*}

\begin{figure}[t]
    \centering
    \includegraphics[width=\linewidth]{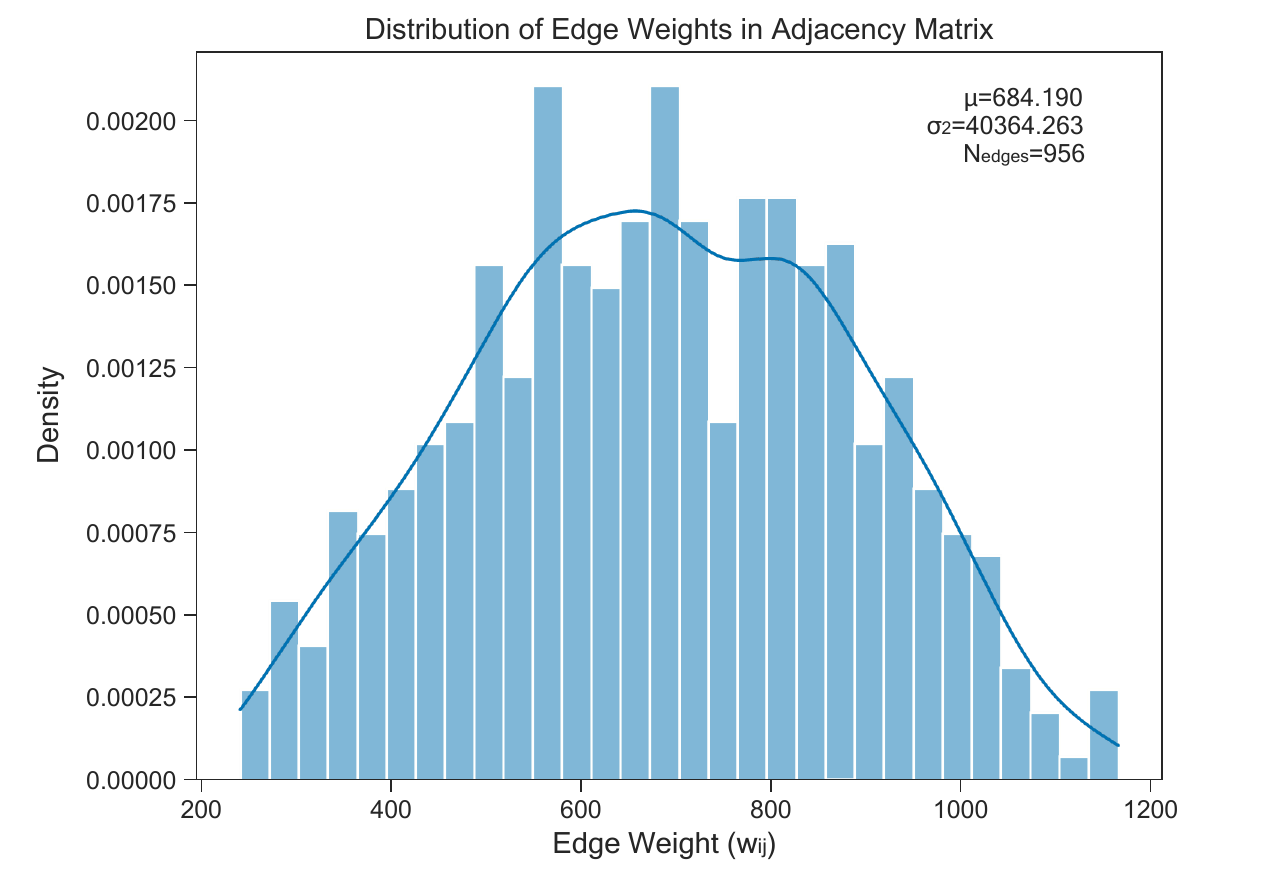}

    \caption{Distribution of edge weights on the GRScenes dataset.}
    \label{sfig:edge_dis}

\end{figure}

Based on the detailed ablation results presented in Table~\ref{stab:ravel_ablation}, the analysis focuses on the specific contributions of individual modules across different scene categories and the statistical stability of the proposed method. The scenario-specific breakdown reveals that the impact of the Minimum Spanning Tree (MSF) strategy is particularly pronounced in unstructured environments. In Home settings, removing the MSF component causes a drastic reduction in Recall, dropping from $0.7625$ to $0.3521$. This sharp decline indicates that in residential layouts, which often feature irregular pathways and visual occlusions, the MSF mechanism is critical for maintaining graph connectivity and preventing the fragmentation of the topological map. Conversely, in Commercial environments, the absence of both the MSF and the Adaptive Verification Parameter (AVP) leads to a significant deterioration in Precision to $0.2728$. This suggests that in commercial spaces characterized by repetitive textures and similar visual features, the adaptive pruning and structural constraints are essential for filtering out false positive connections that simple nearest-neighbor approaches fail to distinguish.Regarding experimental stability, the standard deviations provide insight into the robustness of the method against environmental variance. The data shows that Home scenes generally induce higher performance variability compared to Commercial scenes, as evidenced by the higher standard deviation in Precision for the full method in Home settings ($\pm 0.1291$) versus Commercial settings ($\pm 0.0886$). This disparity reflects the inherent diversity and complexity of residential interiors. Despite this challenge, the proposed method achieves a remarkably low standard deviation for Accuracy across both sub-categories, ranging from $\pm 0.0011$ to $\pm 0.0016$. This consistency demonstrates that the full model effectively mitigates the instability observed in the ablation variants and the baseline Top-k ANN, offering a reliable topological construction regardless of the specific scene characteristics.

\begin{figure}[t]
    \centering
    \includegraphics[width=\linewidth]{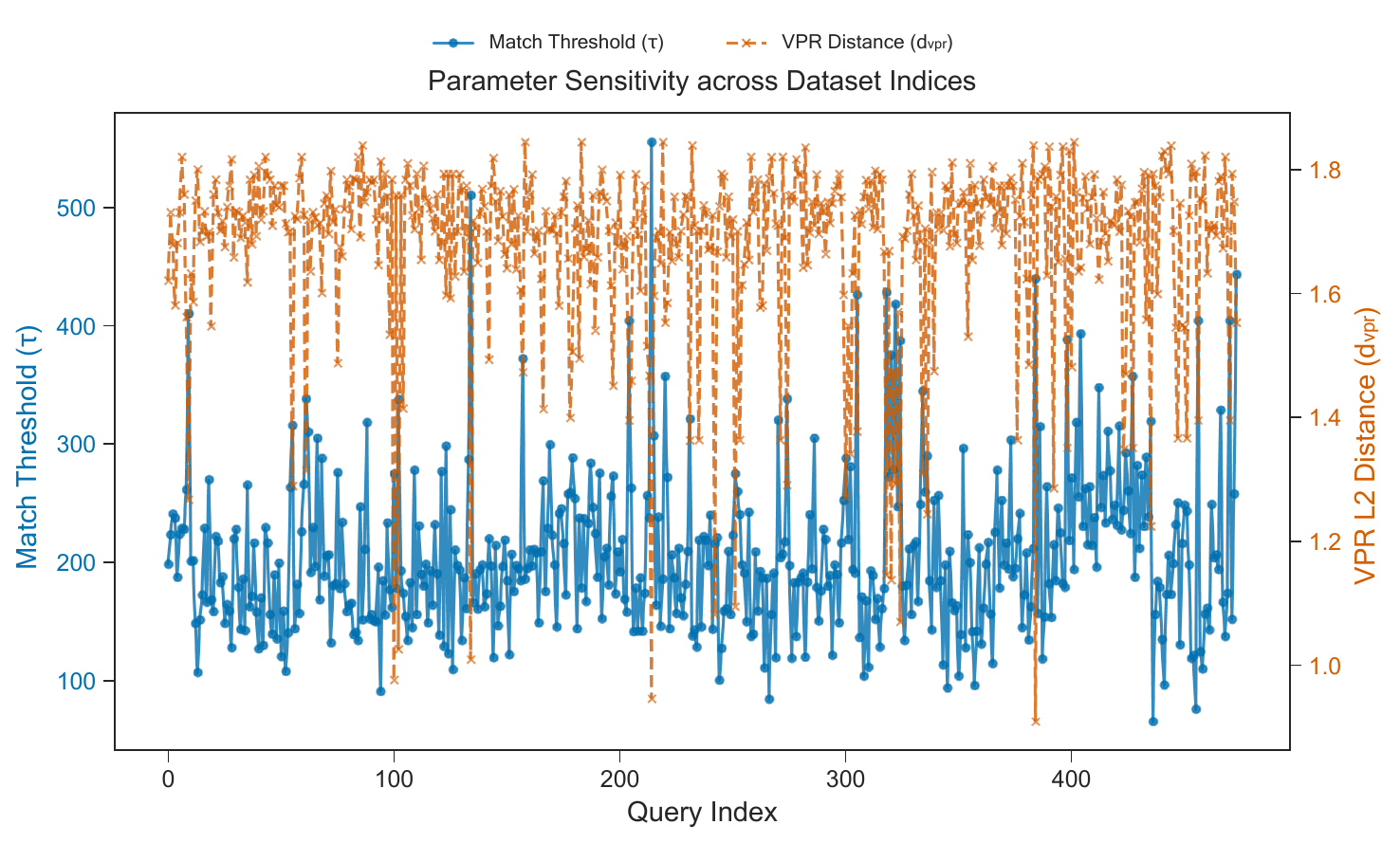}

    \caption{Sensitivity analysis of anchor selection across dataset indices.}
    \label{sfig:param_stability}

\end{figure}

\noindent\textbf{Adaptive Parameter Sensitivity Analysis.}
Based on the provided method description and the statistical visualizations from the GRScenes dataset, the following analysis validates the rationality and effectiveness of the proposed data-driven calibration strategy.

\textit{Validation of Adaptive Thresholding Necessity.}
The distribution of match thresholds ($\tau$) presented in Figure~\ref{sfig:param_dis} (Left) and the sensitivity analysis in Figure~\ref{sfig:param_stability} provide strong empirical evidence for the necessity of an adaptive approach over fixed parameters. The match threshold $\tau$ exhibits a significant spread with a range of $490.00$ and a standard deviation of approximately $4498$ (variance), resulting in a broad Gaussian-like distribution centered around $\mu=205.40$. This high variance indicates that the optimal separation point between inliers and outliers fluctuates drastically depending on the specific visual distinctiveness of the scene. Furthermore, Figure~\ref{sfig:param_stability} illustrates that $\tau$ is not static but varies dynamically across query indices, reacting to the specific appearance properties of the anchor pairs ($q_0, q_f$). A fixed manual threshold would inevitably fail to capture this heterogeneity, leading to either false negatives in texture-poor regions or false positives in repetitive environments. The proposed method, by deriving $\tau$ from the clustering of pooled match counts, effectively centers the threshold for each specific context, ensuring robust feature matching performance.

\textit{Effectiveness of Distance-Based Candidate Selection.}
The distribution of VPR distances ($d_{\text{VPR}}$) in Figure~\ref{sfig:param_dis} (Right) demonstrates the efficacy of the cluster-based estimation in defining a valid search radius. The distribution is heavily skewed towards higher distance values ($\mu=1.68$), with a tight concentration near the upper bound of the range ($0.935$). This statistical behavior aligns with the method's design, where $d_{\text{VPR}}$ is determined by the maximum distance within the highest-confidence cluster ($S_h$). By adapting this boundary, the system maximizes the retrieval of potential loop closure candidates without imposing an arbitrary "top-k" limit. The clear peak in the distribution suggests that the method consistently identifies a "geometric horizon" for verification, successfully distinguishing between the embedding space of valid neighbors and that of unrelated locations.

\textit{Robustness of Graph Connectivity.}
The ultimate validation of the calibration strategy is reflected in the distribution of edge weights shown in Figure~\ref{sfig:edge_dis}. The resulting adjacency matrix weights follow a well-formed normal distribution with a high mean ($\mu=684.19$) and substantial density in the $500-900$ range. This indicates that the calibrated thresholds ($\tau$ and $d_{\text{VPR}}$) successfully filter out low-confidence noise while preserving strong connections. The absence of a significant low-weight tail or bimodal fragmentation suggests that the K-Means clustering logic effectively separates the high-confidence cluster ($S_h$) from the noise ($S_l$). Consequently, the generated topological graph maintains high structural integrity, characterized by strong, verifiable edges rather than weak, ambiguous links. This confirms that the one-shot calibration mechanism functions as an effective filter, translating raw visual data into a high-quality topological representation.

\begin{figure*}[t]
    \centering
    \includegraphics[width=\linewidth]{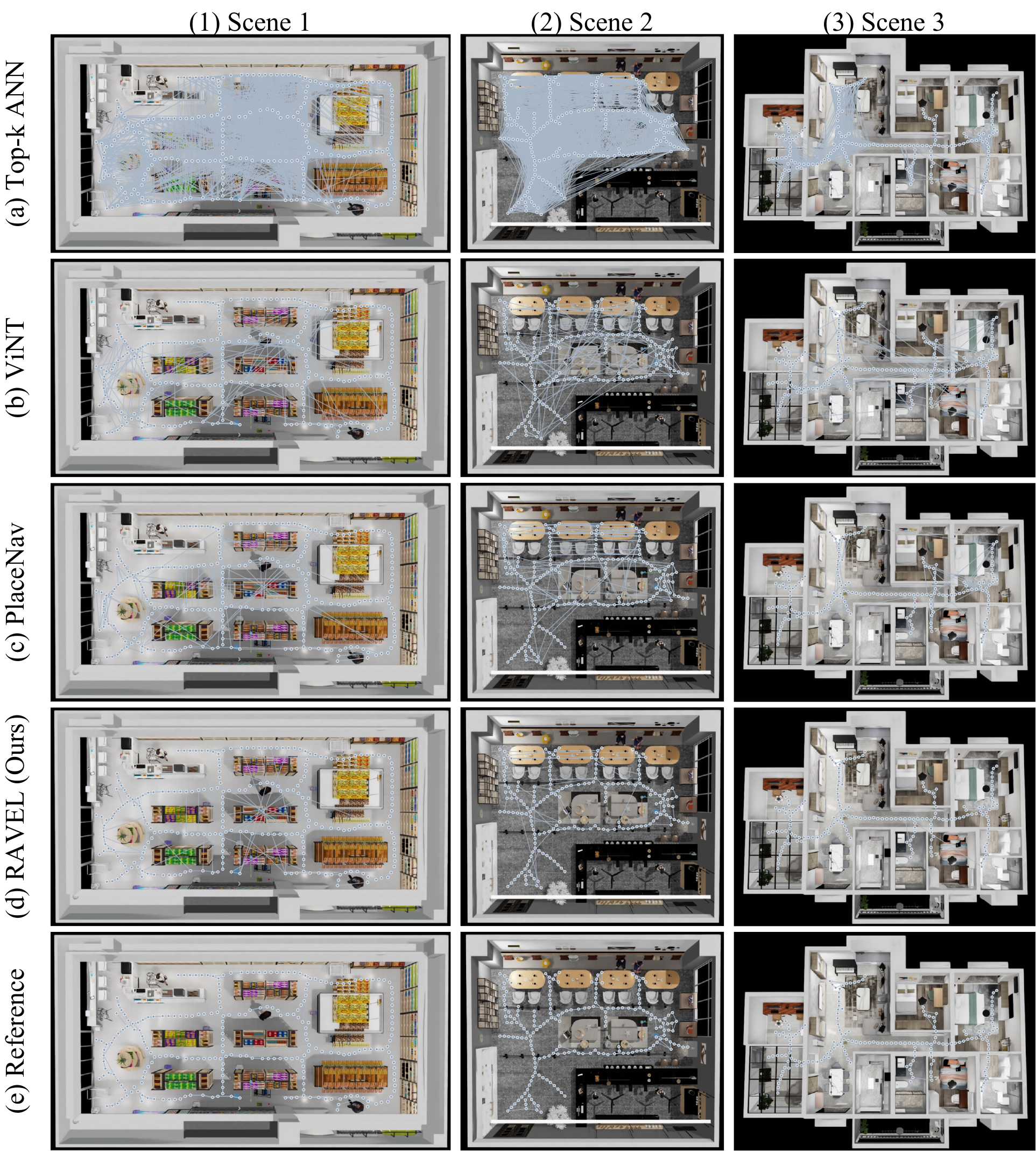}

    \caption{A visualization diagram of topological graphs constructed from unordered images.}
    \label{sfig:topo_vis}

\end{figure*}

\noindent\textbf{Qualitative Analysis of Topological Graph Construction.}
To qualitatively assess the structural integrity and navigability of the generated maps, we visualized the topological graphs constructed by different methods across three distinct scenes. Figure~\ref{sfig:topo_vis} presents a comparative view that contrasts the proposed RAVEL framework against feature-based, temporal-learning-based, and global-retrieval-based baselines, using a manually verified graph as the Reference ground truth.

The visual comparison reveals significant disparities in graph quality, directly highlighting the limitations of heuristic edge filtering. The Top-k ANN approach, which relies solely on the density of LightGlue feature matches, exhibits severe over-connection. As observed in row (a), it generates a chaotic network of edges that frequently penetrate obstacles and walls, particularly in the texture-repetitive environment of Scene 1. This confirms that relying on raw matching counts without rigorous geometric verification fails to distinguish adjacent nodes from visually similar but spatially distant locations. Similarly, the ViNT and PlaceNav methods, while slightly more structured, still suffer from substantial noise. In the open layout of Scene 2 and the multi-room layout of Scene 3, these baselines frequently establish spurious links between disjoint areas, resulting in a "spiderweb" structure that violates the physical constraints of the environment.

In contrast, the proposed RAVEL method produces a topological structure that aligns closely with the Reference ground truth. This superior performance is attributable to two specific components of the RAVEL pipeline. First, the data-driven calibration effectively determines an adaptive inlier threshold $\tau$, which rigorously filters out the weak geometric matches that cause "wall-crossing" edges in the Top-k ANN baseline. Second, and most critically, the two-stage pruning strategy ensures structural cleanliness. By extracting the Maximum Spanning Forest (MSF), RAVEL creates a strong, acyclic skeletal backbone that eliminates the redundant noise seen in Scene 2. Subsequently, the strong-loop reinsertion mechanism selectively restores only the high-confidence cycles necessary for connectivity. This combination allows RAVEL to respect room separation in Scene 3 while maintaining valid traversability through doorways, resulting in a sparse, geometrically consistent, and topologically faithful graph.

\begin{figure*}[t]
    \centering
    \includegraphics[width=\linewidth]{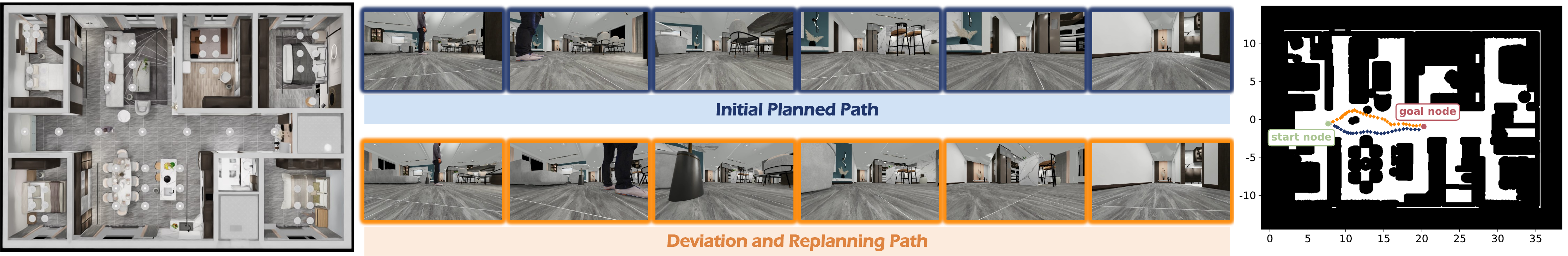}

    \caption{Examples of navigation visualization in the \textit{home} scenario.}
    \label{sfig:replanning_home}

\end{figure*}

\begin{figure*}[t]
    \centering
    \includegraphics[width=\linewidth]{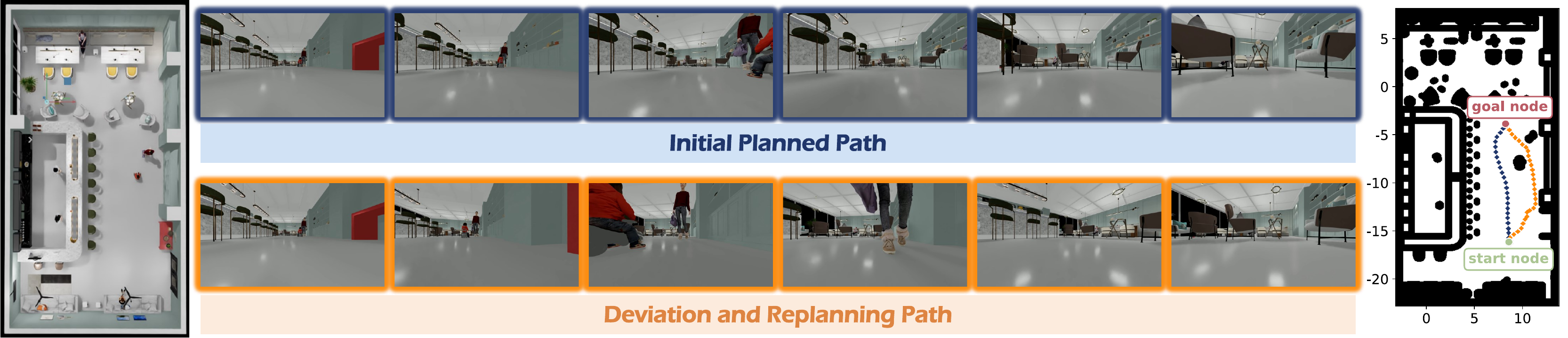}

    \caption{Examples of navigation visualization in the \textit{commercial} scenario.}
    \label{sfig:replanning_commercial}

\end{figure*}

\vspace{8pt}
\noindent\textbf{Localization Performance.}
To evaluate the robustness of the proposed system under realistic navigation challenges, we conducted simulated global localization experiments across multiple time-series datasets. By systematically applying image processing perturbations such as viewpoint modifications and noise injection, we mimicked the sensory imperfections encountered during active robot motion. Table~\ref{stab:localization_accuracy} presents the quantitative comparison of localization accuracy under these varying conditions, offering insights into the stability of different topological localization methods.

The results demonstrate that the proposed BPL method achieves superior generalization across all evaluated benchmarks, yielding the highest overall average accuracy of $0.9304$. This performance significantly outperforms the competing baselines, particularly in scenarios involving complex visual degradations. A critical limitation observed in baseline methods, specifically MegaLoc, is a marked sensitivity to high-frequency noise. While MegaLoc performs competitively in clean rotation scenarios on the GoStanford dataset, its accuracy deteriorates rapidly when Gaussian or Poisson noise is introduced, dropping by nearly $30\%$ in datasets such as SACSoN and SCAND. In contrast, BPL exhibits exceptional resilience to these signal corruptions. The method maintains consistently high accuracy levels even under "Rot + Gauss" and "Rot + Poisson" conditions, suggesting that its feature representation is effective at suppressing sensor noise while retaining structurally relevant topological information.

Furthermore, the "Rot + Crop" condition serves as a stress test for partial observability and drastic viewpoint shifts. In this regime, methods like ViNT and JIST experience substantial performance degradation, particularly on the SACSoN and SCAND datasets where ViNT's accuracy falls significantly. The proposed BPL framework, however, consistently achieves accuracy exceeding $0.92$ in these occlusion-heavy settings. This indicates that BPL successfully leverages global context to localize the agent even when significant portions of the visual field are missing or altered. Consequently, these findings confirm that BPL provides a reliable localization solution that is not only accurate in ideal conditions but remains robust against the unpredictable visual variations inherent in real-world robotic navigation.

\vspace{8pt}
\noindent\textbf{Navigation Performance.}
To qualitatively evaluate the practical applicability of the constructed topological graphs, we visualized the navigation process within realistic simulation environments. This evaluation encompasses the full navigation pipeline, ranging from graph construction using unordered images to start-goal localization and path planning executed by a local planner. A particular focus is placed on the system's dynamic response to execution errors, where the Belief Propagation Localization (BPL) module identifies deviations from the optimal trajectory and triggers topological replanning to ensure successful navigation.

Figures~\ref{sfig:replanning_home} and~\ref{sfig:replanning_commercial} illustrate the navigation performance in Home and Commercial settings, respectively. In both scenarios, the blue trajectories represent the initial optimal paths computed on the topological graph, while the orange trajectories depict instances where the agent deviates from the intended route due to control noise or local obstacle avoidance. The visual data presented in the "Deviation and Replanning" rows highlights that during these excursions, the robot encounters viewpoints that differ significantly from the initial plan. Despite these visual discrepancies, the BPL module successfully detects the mismatch between the expected and actual states, prompting the system to generate a new topological path.

The successful recovery in both distinct environments demonstrates the robustness of the proposed framework. In the clutter-heavy Home scenario, the robot effectively navigates around central obstacles after deviating, proving that the constructed graph maintains sufficient connectivity even in narrow, irregular spaces. Similarly, in the texture-repetitive Commercial scenario, the system corrects a significant lateral drift, indicating that the localization module is resilient to the perceptual aliasing common in large open halls. These results confirm that the topological graphs constructed from unordered images provide a reliable spatial representation that, when coupled with BPL, allows for autonomous self-correction and consistent goal achievement under imperfect execution conditions.

\vspace{8pt}
\noindent\textbf{Summary of Experimental Results.}
In this section, we have provided extended experimental results to evaluate the performance of ULVN across mapping, localization, and navigation tasks. ULVN outperforms baseline methods in all tasks, demonstrating its robustness and effectiveness in constructing reliable topological maps, localizing within those maps, and navigating through complex environments. The results from both simulated and real-world environments showcase the practical potential of ULVN for autonomous navigation in unstructured spaces.

\end{document}